\documentclass[letter,11pt]{article}

\usepackage[utf8]{inputenc}
\usepackage[T1]{fontenc}

\usepackage[a4paper,margin=1in]{geometry}

\usepackage{color}
\usepackage[colorlinks=true,linkcolor=blue,citecolor=blue]{hyperref}
\usepackage{amsmath, amssymb, amsthm}
\usepackage{mathtools}
\usepackage{graphics}
\usepackage{pifont}
\usepackage{tikz}
\usepackage{bbm, bm}
\usepackage{lipsum}
\DeclareMathOperator*{\argmax}{arg\,max}

\usetikzlibrary{arrows.meta}
\usepackage{environ}
\usepackage{framed}
\usepackage{url}
\usepackage[framemethod=tikz]{mdframed}
\usepackage{appendix}
\usepackage{graphicx}
\usepackage{tcolorbox}
\allowdisplaybreaks[1]
\usepackage{nicefrac}
\usepackage{natbib}
\usepackage{thm-restate}
\usepackage[noabbrev,capitalize,nameinlink]{cleveref}
\crefname{equation}{}{}

\newcommand\remove[1]{}

\newtheorem{lemma}{Lemma}[section]
\newtheorem*{lemma*}{Lemma}
\newtheorem{theorem}[lemma]{Theorem}

\newtheorem*{corollary*}{Corollary}

\newtheorem{proposition}{Proposition}
\newtheorem{remark}{Remark}

\newtheorem*{theorem*}{Theorem}
\newtheorem*{inducthyp*}{Inductive Hypothesis}
\newtheorem*{definition*}{Definition}
\newtheorem{definition}[lemma]{Definition}

\newtheorem*{rem*}{Remark}

\newcommand\R{\mathbb{R}}
\newcommand\N{\mathbb{N}}

\newcommand\E{\mathbb{E}}
\newcommand{\bitset}{\{0,1\}}

\renewcommand{\H}{\mathcal{H}}

\renewcommand{\hat}{\widehat}
\renewcommand{\tilde}{\widetilde}

\DeclareFontFamily{U}{mathb}{\hyphenchar\font45}
\DeclareFontShape{U}{mathb}{m}{n}{<5> <6> <7> <8> <9> <10> gen * mathb
<10.95> mathb10 <12> <14.4> <17.28> <20.74> <24.88> mathb12}{}
\DeclareSymbolFont{mathb}{U}{mathb}{m}{n}
\DeclareMathSymbol{\rcirclearrow}{\mathbin}{mathb}{'367}

\foreach \x in {a,...,z}{%
	\expandafter\xdef\csname b\x\endcsname{\noexpand\mathbf{\x}}
}

\foreach \x in {A,...,Z}{%
	\expandafter\xdef\csname m\x\endcsname{\noexpand\mathbf{\x}}
}

\foreach \x in {A,...,Z}{%
	\expandafter\xdef\csname c\x\endcsname{\noexpand\mathcal{\x}}
}

\newif\ifrandom
\randomtrue

\newcommand{\V}{\mathcal{V}}
\newcommand{\U}{\mathcal{U}}

\newcommand{\todolater}[1]{}

\DeclareUnicodeCharacter{2113}{$\ell$}

\usepackage{multirow}

\newcommand{\lr}[1]{\mathopen{}\left(#1\right)}

\newcommand{\lrbra}[1]{\mathopen{}\left[#1\right]}

\newcommand{\lrset}[1]{\mathopen{}\left\{#1\right\}}

\newcommand{\lrabs}[1]{\mathopen{}\left|#1\right|}

\newcommand{\VC}{\mathrm{VC}}
\newcommand{\LD}{\mathrm{LDim}}

\newcommand{\Sdim}{\mathfrak{s}_{\emptyset}}
\newcommand{\sdsolver}{\mathrm{SDSolver}}
\newcommand{\SDdim}{\mathrm{SDdim}}
\newcommand{\Ldim}{\mathrm{LDim}}

\newcommand{\1}{\mathbbm{1}}
\newcommand{\Mopt}{M_{\mathrm{optimal}}}
\newcommand{\Ropt}{R_{\mathrm{optimal}}}
\newcommand{\X}{\mathcal{X}}
\usepackage{algorithm}
\usepackage{algpseudocode}

\title{Self-Directed Apple Tasting}

\author{
    Tanmay Devale \thanks{Department of Computer Science, Purdue University; \texttt{tdevale@purdue.edu}} 
    \and  Reed Phillips \thanks{Department of Computer Science, Purdue University; \texttt{phill289@purdue.edu}}
    \and  Arvind Ramaswami\thanks{Department of Computer Science, Purdue University; \texttt{ramaswa4@purdue.edu}}
}
\usepackage{times}

\begin{document}

\maketitle

\begin{abstract}
  We study the classical apple-tasting setting, a partial feedback model in which the learner observes the true label only when it predicts $1$. A combinatorial characterization for the rates of this problem in the online adversarial setting has been studied by \cite{raman2024appletastingcombinatorialdimensions}.
  We study the \emph{self-directed} setting, in which the learner can adaptively choose the order of the points it predicts on.
  In the realizable setting, we establish a tetrachotomy on the expected minimax number of mistakes made by any learner. 
  Depending on the VC dimension, the $\emptyset$-centered star number, and the self-directed dimension, we show that the rates could possibly be $\Theta(1)$, $O(\log n) \cap \omega(1)$, $\tilde \Theta(\sqrt{n})$, or $\Theta(n)$. 
  In the agnostic setting, the minimax regret is $\tilde \Theta(\sqrt{n})$ or $\Theta(n)$,  depending on the VC dimension.

\end{abstract}

\section{Introduction}
Imagine visiting a fruit vendor to buy apples: you pick one, taste it, and learn whether it is good; you then pick another, taste it, and so on. 
We call this \emph{self-directed online classification under
apple-tasting feedback}, an analogue of the model introduced by \cite{Helmbold2000}.
In this model, the learner observes the correct label only when it predicts $1$ (i.e. eats the apple), in contrast to standard online learning, where the correct label is revealed after every prediction.

We believe that this self-directed formulation is arguably the natural way to model literal apple tasting: the defining feature of the problem is that the learner controls \emph{which} point it exposes itself to risk on, and self-direction simply takes this control to its logical conclusion by letting the learner choose the point itself, rather than having it imposed by an adversary. 
Prior work on apple tasting
\citep{raman2024appletastingcombinatorialdimensions,chase2025deterministicappletasting} instead studies the setting in which the point in each round is chosen by the environment or adversary, and the learner only controls its prediction. 
While this adversarial formulation is a natural starting point for worst-case analysis, it discards exactly the element of agency that makes apple tasting a distinctive feedback model to begin with --- after all, no one is forced to buy a particular apple at the market. 

A more practical instance of this problem can be seen in the hiring process at a company. They first review applications and then decide which candidates to interview. 
As interviewing is more informative but more resource-intensive than simply reading applications, the company might want to adaptively decide which candidates to interview in order to learn what features in applications are most desirable.

Formally, self-directed online classification with apple-tasting feedback is a repeated game between an \emph{adversary} and a \emph{learner} played over $n$ rounds. 
At the start of the game, the adversary provides a set of $n$ points. In round $i$, the learner selects one of these points $x_i$ and makes a prediction $\hat y_i \in \bitset$. 
The adversary then assigns a label $y_i \in \bitset$, unknown to the learner, and the learner incurs loss $\1[\hat y_i \neq y_i]$. 
If $\hat y_i = 1$, the adversary reveals the true label $y_i$; otherwise, the learner receives no feedback. We allow both the learner and the adversary to use randomized strategies. In most cases, however, we show that deterministic strategies would give essentially the same results.

We study both the \emph{realizable} and the \emph{agnostic} settings in our paper. We adopt the perspective of \cite{raman2024appletastingcombinatorialdimensions} and identify the combinatorial dimensions that govern the minimax number of mistakes and regret\footnote{Number of mistakes compared to the best concept in the concept class.} respectively. Our main contributions are as follows:
\begin{itemize}
    \item The Vapnik-Chervonenkis (VC) Dimension \citep{Vapnik1971} characterizes the minimax expected regret in the agnostic setting as $\tilde \Theta \lr{\sqrt{n \cdot \VC(\H)}}$.
    \item On the other hand, VC dimension alone is not enough to characterize the realizable setting. 
    There are easier regimes characterized by the $\emptyset$-centered star number of $\H$ \citep{hanneke2024starnumbereluderdimension}, denoted $\Sdim(\H)$, and the self-directed dimension of \citep{devulapalli2024dimensionselfdirectedlearning} denoted by $\SDdim(\H)$.
    \item Consequently, in the realizable setting, we obtain a tetrachotomy on the mistake bound. 
    (i) $\Theta_\H(1)$ when $\Sdim(\H) < \infty$, $\SDdim(\H)<\infty$;
    (ii) $O_\H(\log n) \cap \omega_\H(1)$ when $\Sdim(\H)<\infty$ but $\SDdim(\H)=\infty$;
    (iii) $\tilde\Theta_\H(\sqrt{n})$ when $\VC(\H) < \infty$ but $\Sdim(\H) = \infty$;
    (iv) $\Theta(n)$ when $\VC(\H) = \infty$.
\end{itemize}
We additionally establish several auxiliary results.
\begin{itemize}
    \item We show that for any $d \in \N$, there exists a concept class $\H_d$ such that $\VC(\H_d)=d$ and the optimal number of mistakes made by any algorithm is $\Omega(\sqrt{nd})$.
    \item In the realizable setting, if the adversary may choose labels adaptively, our upper and lower bounds on the mistake bound differ by a factor of $\sqrt{\log n}$. 
    This gap vanishes when the adversary must instead commit to an unknown true concept before the
    game begins.
    \item It is straightforward to find concept classes that fit in cases (i), (iii), and (iv) of our tetrachotomy. We show that case (ii), where $\Sdim(\H) < \infty$ and $\SDdim(\H) = \infty$, is also nonempty by adapting a construction of \citet{kuhlmann1999teachingintersectionclosed}.
    \item Finally, we prove a purely combinatorial identity that may be of independent
    interest. For any concept class $\H$, letting $\U(\H)$ denote its union closure, we have $\Sdim(\H)=\Sdim(\U(\H))=\VC(\U(\H))$.
\end{itemize}

\section{Literature Review}

\textbf{Apple Tasting:} Apple tasting is a partial feedback model. It is a special case of partial monitoring (\cite{bartok2014partial}) and of sequential prediction with graph feedback (\cite{alon2015beyond}). Online learning with apple-tasting feedback was initially studied in \cite{Helmbold2000}, via randomized algorithms and an analysis of expected regret. \cite{raman2024appletastingcombinatorialdimensions} provide a combinatorial characterization for online apple tasting, via the Littlestone dimension $\LD(\H)$ in the agnostic case, and show an upper bound of $O(\sqrt{\Ldim(\H)T\log T})$ on the expected regret. They provide a tighter combinatorial characterization using the \textit{effective width} in the realizable case where they show that the minimax mistake bound exhibits a trichotomy: $\Theta(1)$, $\Theta(\sqrt T)$, or $\Theta(T)$. Our tetrachotomy (Theorem \ref{thm:rates}) is a self-directed counterpart of this result, where the rates are instead governed by the VC dimension, $\emptyset$-centered star number, and the self-directed dimension.

\cite{chase2025deterministicappletasting} study deterministic algorithms and show a $\sqrt{\log T}$ gap between randomized and deterministic learners in the realizable case. This gap could also appear in our setting; all of our upper bounds in Section \ref{sec:upper} are achieved by deterministic learners, while our lower bounds in Section \ref{sec:lower} hold against randomized ones.

\textbf{Self-Directed Learning:} Self-directed learning was studied for special classes in \cite{goldman1994power}. In more generality, \cite{devulapalli2024dimensionselfdirectedlearning} provide a dimension $\mathrm{SDdim}$ to characterize the mistake bound for self-directed learning. Our work considers self-directed learning with apple-tasting rather than full-information feedback. \cite{bendavid1995self} and \cite{kuhlmann1999teachingintersectionclosed} provide separations between the number of mistakes in self-directed learning and offline learning. We use the intersection-closed property of \cite{kuhlmann1999teachingintersectionclosed}'s construction in Theorem \ref{thm:intersectionclosed}.

\textbf{The Star Number:} \cite{hanneke15a} introduced the star number $\mathfrak{s}(\H)$ to characterize the minimax label complexity in active learning under low noise regimes under the noise model of \cite{tsybakov2004optimal}. It has since appeared in several other roles: \cite{wiener2015compression} provide a version space compression scheme of size $\mathfrak{s}(\H)$, and \cite{hanneke2016refined} analyze its role in the sample complexity of empirical risk minimization (ERM). \cite{bousquet2020proper} study a related variant, the \textit{hollow star number}, in connection with the Helly number and an optimal SVM bound. \cite{hanneke2024starnumbereluderdimension} collects elementary observations about the star number, such as equivalence with the VC dimension of the version space class. 

Our work uses the $\emptyset$-centered star number, introduced by \cite{hanneke2024starnumbereluderdimension}, as one of the parameters to analyze mistake bounds for self-directed apple tasting. 
Intuitively, star number measures the worst-case deviation from a particular central concept. 
Apple tasting's asymmetry makes the zero concept especially important.
Using it as the center yields the $\emptyset$-centered star number.

\textbf{Prediction with Expert Advice:} Prediction with expert advice \citep{cesa2006prediction,vovk1990aggregating,vovk1995game} is a sequential framework for making predictions. A learner receives predictions from $N$ experts in each of $T$ rounds and must make few mistakes relative to the best-performing expert. Multiplicative weights gives a regret bound of $\sqrt{T \log N}$. Tighter first- and second-order bounds were studied, such as the second-order bounds in \cite{cesa2007improved}. 
We specifically use the expert advice algorithm with an apple-tasting variant EXP4.AT of \cite{raman2024appletastingcombinatorialdimensions}.

\textbf{Other tools:} Two results are used directly. 
We use the one-inclusion graph predictor of \cite{Haussler1994}, whose leave-one-out guarantee bounds the expected error of predicting an unseen point by $\VC(\H)/|R|$ given $|R|$ training points. 
We also use progress on the union-closed set conjecture, which guarantees that some point must be labeled $1$ by a constant fraction of a union-closed concept class. See Proposition \ref{prop:union-closed} and \cite{bruhn2013journeyunionclosedsetsconjecture} for a survey of the conjecture.

\section{Formal Setup and Definitions}
In this section, we formally present the setup and definitions used throughout the paper. 

Let $\mathcal X$ denote an arbitrary non-empty set, referred to as the \emph{instance space}, and $\mathcal Y = \{0,1\}$ the \emph{label space}. 
Let $\H$ denote an arbitrary non-empty set of concepts $h: \mathcal X \to \mathcal Y$, called the \emph{concept class}.

Our framework, self-directed online learning under apple-tasting feedback, combines two existing models. 
As in the self-directed online learning model of \citet{bendavid1995self}, the learner chooses which point to query in each round. 
As in the apple-tasting model of \citet{Helmbold2000}, feedback is partial: the learner observes the true label when it predicts $1$, and receives no feedback when it predicts $0$.

Unlike in full-information settings, the adversary may benefit by presenting the learner with multiple copies of the same point. Conversely, the learner may benefit from labeling different copies of the same point differently. We therefore allow the learner to choose points from a multiset $S$ and use the notation $S \subseteq \X$ to mean that every $x \in S$ is also in $\X$. 

\subsection{Self-directed online learning under apple-tasting feedback setting}
We describe the learning protocol. 
Given a non-empty concept class $\H$ and a finite multiset $S \subseteq \mathcal X$, one round of self-directed online learning under apple-tasting feedback proceeds as follows.
\begin{enumerate}
    \item The learner chooses some $x \in S$.
    \item The learner predicts a label $\hat y \in \mathcal Y$ for $x$.
    \item If $\hat y = 1$, the learner receives the true label $y \in \mathcal Y$ from the
    environment. If $\hat y = 0$, the learner receives no feedback.
\end{enumerate}
Rounds proceed in this fashion until the learner has labeled every point in $S$. For brevity we say the learner \emph{eats} a point $x$ if they predict $1$ on it.
\subsection{Definitions}

Because feedback is partial under apple tasting, the histories available to the learner and to the adversary differ. When the learner predicts $0$, no true label is revealed, so the learner records the symbol $\star$ in its history in place of a label.
\begin{definition} 
Given a finite multiset $S \subseteq \mathcal X$, we formalize an adaptive learner as a pair
of functions
\begin{align*}
    \mathsf{Lrn}_x:(S \times (\mathcal Y \cup \{\star\}))^* \rightarrow \Pi(S), \quad \mathsf{Lrn}_y:(S \times (\mathcal Y \cup \{\star\}))^* \times S \rightarrow \Pi(\mathcal{Y}),
\end{align*}
Where $\Pi(S)$ and $\Pi(\mathcal{Y})$ are the sets of distributions over $S$ and $\mathcal{Y}$, respectively. The adversary is a function 
\begin{align*}
    \mathsf{Adv}:(S \times \mathcal Y)^* \times S \rightarrow \Pi(\mathcal{Y})
\end{align*}
\end{definition}
In round $i$, $\mathsf{Lrn}_x$ is used in the first step to choose a distribution from which the queried instance $x_i$ is sampled. $\mathsf{Lrn}_y$ is used in the second step to determine the prediction $\widehat{y}_i$ of the learner. 
$\mathsf{Adv}$ is used in the third step to determine the correct label $y_i \in \mathcal Y$. 

\begin{definition}
We say a labeled sequence $S_n = ((x_1,y_1), \ldots, (x_n,y_n))$ is \emph{realizable}
under a concept class $\H$ if
\begin{equation}
    \min_{h \in \H} \sum_{i=1}^n \1[h(x_i) \neq y_i] = 0.
\end{equation}
\end{definition}
Let $F_\mathsf{Lrn} \in (S \times (\mathcal Y\cup\{\star\}))^*$ be the learner's history.
Let $F_{\mathsf{Adv}} \in (S \times \mathcal Y)^*$ be adversary's history.
We say an adversary is \emph{consistent} with $\H$ if every history $F_\mathsf{Adv}$ that could arise under their strategy is realizable under $\H$.

We now define the optimal mistake bound against consistent adversaries, commonly known as the \textbf{\emph{realizable setting}}.
\begin{definition}
\label{def:realizableMistakeBound}
For a learner $\mathsf{Lrn}$ and an adversary $\mathsf{Adv}$, let $F^{(i)}_\mathsf{Lrn}$ and $ F^{(i)}_\mathsf{Adv}$ denote the respective histories at the start of round $i$ and let
\begin{equation*}
    x_i \sim \mathsf{Lrn}_x\lr{F^{(i)}_\mathsf{Lrn}} \qquad 
    y_i \sim \mathsf{Adv}\lr{F^{(i)}_\mathsf{Adv},x_i} \qquad
    \hat y_i \sim \mathsf{Lrn}_y\lr{F^{(i)}_\mathsf{Lrn},x_i}
\end{equation*} 
denote, the instance selected by the learner, the label chosen by the
adversary, and the label predicted by the learner in round $i$ respectively.

Let a finite multiset $S \subseteq \mathcal X$ of size $n$ be given to the learner. Let the game last for $m$ rounds. We define the expected number of mistakes the learner makes against the adversary by
\begin{equation}
    M(S,\mathsf{Lrn},\mathsf{Adv}) = \E\lrbra{\sum_{i=1}^{m} \1\lrbra{\hat y_i \neq y_i}}
\end{equation}

The learner's history evolves as 
\begin{equation*}
    F^{(i+1)}_\mathsf{Lrn} =
    \begin{cases}
        F^{(i)}_\mathsf{Lrn} \circ (x_i, y_i) & \text{if } \hat y_i = 1, \\
        F^{(i)}_\mathsf{Lrn} \circ (x_i, \star) & \text{if } \hat y_i = 0
    \end{cases}  
\end{equation*}
and the adversary's history evolves as
\begin{equation*}
    F^{(i+1)}_\mathsf{Adv}=F^{(i)}_\mathsf{Adv}\circ (x_i,y_i)
\end{equation*}
where $F^{(0)}_\mathsf{Lrn}=F^{(0)}_\mathsf{Adv}=\emptyset.$

We define the optimal mistake bound of $\H$ against an adversary providing apple-tasting feedback on $S$ by
\begin{equation}
    M(\H,S)=\inf_\mathsf{Lrn} \sup_\mathsf{Adv} M(S,\mathsf{Lrn},\mathsf{Adv})
\end{equation}
where the supremum is taken over all adversaries which are consistent with $\H$.

Finally, we define the optimal mistake bound over over all finite submultisets of $\mathcal X$ of size at most $n$ by
\begin{equation}
    \Mopt(\H,n)= \sup_{S \subseteq \mathcal X, |S|\le n} M(\H,S)
\end{equation}
\end{definition}
\begin{definition}
    We say that a concept class $\H$ is \emph{learnable} under the realizable setting of self-directed online learning with apple-tasting feedback (or learnable, in short) if $M_{\mathrm{optimal}}(\H,n) \in o(n)$\footnote{This is a weaker definition of learnable than used by \cite{devulapalli2024dimensionselfdirectedlearning}; they required $\Mopt(\H, n) \in O_{\H}(1)$. 
    Due to our weaker definition, the gap noted between the self-directed online learning and PAC-setting noted in \cite{devulapalli2024dimensionselfdirectedlearning} disappears.}, and \emph{not learnable} if $M_{\mathrm{optimal}}(\H,n) \in \Omega(n)$.
\end{definition}
If the adversary is not consistent with the concept class then we try to do as well as the best concept in the concept class. This setting is commonly known as a the \textbf{\emph{agnostic setting.}} As a result, we define expected regret as the number of mistakes made by an algorithm in comparison to the best concept in the concept class.
\begin{definition}
    The learner $\mathsf{Lrn}$, the adversary $\mathsf{Adv}$, the history of the learner $F^{(i)}_\mathsf{Lrn}$, the history of the adversary $F^{(i)}_\mathsf{Adv}$, the instance selected by the learner $x_i$, the label chosen by the adversary $y_i$ and the label predicted by the learner $\hat y_i$ in round $i$ are exactly the same as in Definition \ref{def:realizableMistakeBound}.

    For a given finite multiset $S \subseteq \X$ of size $n$, let the game go on for $m$ rounds. We define the expected regret of the learner against the adversary as 
    \begin{equation}
        R(S,\mathsf{Lrn},\mathsf{Adv})=\E\lrbra{\sum_{i=1}^m \1\lrbra{\hat y_i\neq y_i}-\inf_{h \in \H}\sum_{i=1}^m \1\lrbra{h(x_i)\neq y_i}}
    \end{equation}
    We define the optimal expected regret of $\H$ against an adversary providing apple-tasting feedback on $S$ by
    \begin{equation}
        R(\H,S)= \inf_\mathsf{Lrn} \sup_\mathsf{Adv} R(S,\mathsf{Lrn},\mathsf{Adv})
    \end{equation}
    where the supremum is taken over all adversaries.

    Finally, we define the optimal mistake bound over all finite submultisets of $\X$ of size at most $n$ by
    \begin{equation}
        \Ropt(\H,n) = \sup_{S \subseteq \X,|S|\le n} R(\H,S)
    \end{equation}
\end{definition}
\begin{definition}
    We say that a concept class $\H$ is learnable under self-directed online learning with apple-tasting feedback in the agnostic setting (or agnostic learnable, in short) if $\Ropt(\H,n) \in o(n)$, and not agnostic learnable if $\Ropt(\H, n) \in \Omega(n)$.\footnote{Since in the agnostic setting the adversary does not have to be consistent, our notion of agnostic learnability is no weaker than realizable learnability. In fact, our results show that they are equivalent.}
\end{definition}
For the sake of completeness, we restate the definition of $\VC$ dimension, $\emptyset$-centered star number, and self-directed dimension. 
\begin{definition}[\textbf{VC dimension }\citealp{Vapnik1971}]
\label{def:vc}
A set $S$ is \emph{VC-shattered} by a concept class $\H$ if the projection of $\H$ onto
$S$, i.e., $\H|_S$, has size $2^{|S|}$. The \emph{VC dimension} of $\H$, denoted
$\VC(\H)$, is the size of the largest set that $\H$ VC-shatters, or $\infty$ if no such
largest set exists.
\end{definition}
\begin{definition}[\textbf{$\emptyset$-centered star number }\citealp{hanneke15a,hanneke2024starnumbereluderdimension}]
\label{def:Sdim}
    A set $S$ is \emph{$\emptyset$-centered star-shattered} by a concept class $\H$ if:
    \begin{itemize}
        \item[(1)] There exists a concept $h \in \H$ such that $h(x) = 0$ for all $x \in S$; and
        \item[(2)] For every $y \in S$, there exists a concept $h \in H$ such that $h(y) = 1$ and $h(x) = 0$ for all $x \in S \setminus\{y\}$.
    \end{itemize}
    The \emph{$\emptyset$-centered star number} of $\H$, denoted $\Sdim(\H)$, is the size of the largest set that $\H$ $\emptyset$-centered star-shatters, or $\infty$ if no such largest set exists.

    For an $\emptyset$-centered star-shattered set, we say that the concept realizing (1) is the \emph{center} and the concepts realizing (2) are the \emph{witnesses}.
\end{definition}
The self-directed dimension is the dimension that characterizes the self-directed online learning learning with full-information setting.
\begin{definition}[\textbf{Self-directed dimension }\citealp{devulapalli2024dimensionselfdirectedlearning}]
\label{def:SDdim}
For a concept class $\H$ on a finite multiset $S$, let $\V \subseteq \H$. We define $M(\V, S)$ by $M_{\mathrm{full}}(\V,S) = 0$ whenever $|\V|_S| \le 1$, and otherwise
\[
    M_{\mathrm{full}}(\V, S) = \min_{x \in S}\ \min_{\hat y \in \bitset}\ \max_{y \in \bitset} \lrbra{\1[\hat y \neq y] + M_{\mathrm{full}}\lr{\V_{x \to y},\ S \setminus \lrset{x}}}.
\]
The recursion terminates since $|S|$ strictly decreases. The \emph{dimension of self-directed learning} of $\H$ is $\SDdim(\H) = \sup_{S} M_{\mathrm{full}}(\H|_S, S)$, the supremum over finite multisets $S$. 

When $\SDdim(\H) < \infty$, \cite{devulapalli2024dimensionselfdirectedlearning} provide a deterministic algorithm that makes at most $\SDdim(\H)$ mistakes on any input. We call this the $\sdsolver$ algorithm and use it as a black box.
\end{definition}
\begin{definition}[Union-closure]
\label{def:union-closure}
    Let $\H$ be a nonempty concept class. For a subset $\mathcal{T} \subseteq \H$, let $h_{\mathcal{T}}$ denote the concept for which $h_{\mathcal{T}}(x) = 1$ if and only if $h(x) = 1$ for some $h \in \mathcal{T}$. Then the union-closure $\U(\H)$ of $\H$ is $\U(\H) = \{h_\mathcal T: \mathcal T \subseteq \H, \mathcal{T} \neq \emptyset\}$.
\end{definition}
Unless stated otherwise, $O,\Omega,\Theta$ hide universal constants which do not depend on any parameter of the problem. Notation such as $\Theta_{\H}$ indicates that the implied constants depend on the concept class $\H$.

\section{Tetrachotomy in the realizable setting}
In this section, we focus on the realizable setting. The following theorem is the main result of this work. It expresses a fundamental \emph{tetrachotomy}: there are exactly four possibilities for optimal learning rates.

\begin{theorem}
\label{thm:rates}
    For every concept class $\H$ exactly one of the following holds:
    \begin{itemize}
        \item $\H$ is not learnable, i.e. $\Mopt(\H,n) \in \Omega(n)$.
        \item $\H$ is learnable with  $\Mopt(\H,n) \in  \widetilde{\Theta}_{\H}(\sqrt{n})$.
        \item $\H$ is learnable with $\Mopt(\H,n) \in O_{\H}(\log n) \cap \omega_\H(1)$.
        \item $\H$ is learnable with $\Mopt(\H,n) \in \Theta_\H(1).$
    \end{itemize}
\end{theorem}
A second main result of this work provides a detailed description of which of these four cases any given concept class $\H$ belongs to, by specifying complexity measures to distinguish the cases.

We use the well-known \emph{Vapnik Chervonenkis dimension} denoted $\VC(\H)$ as defined in \cite{Vapnik1971}, the \emph{$\emptyset$-centered star number}, denoted $\Sdim(\H)$ as defined in \cite{hanneke15a}, and the \emph{self-directed dimension} $\SDdim(\H)$ as defined in \cite{devulapalli2024dimensionselfdirectedlearning} to characterize the various cases.

\begin{theorem}
\label{thm:dimensionforrates}
    Every concept class $\H$ falls into one of the following four cases:
    \begin{itemize}
        \item If $\VC(\H)=\infty$ then $\H$ is not learnable, i.e. $\Mopt(\H,n) \in \Omega(n)$.
        \item If $\VC(\H)<\infty$ and $\Sdim(\H)=\infty$ then $\H$ is learnable with $\Mopt(\H,n) \in \widetilde{\Theta}_\H(\sqrt{n})$. 
        \item If $\Sdim(\H)<\infty$ and $\SDdim(\H)=\infty$ then $\H$ is learnable with $\Mopt(\H,n) \in O_\H(\log n) \cap \omega_\H(1)$.
        \item If $\Sdim(\H)<\infty$ and $\SDdim(\H)<\infty$ then $\H$ is learnable with $\Mopt(\H,n) \in \Theta_{\H}(1)$.
    \end{itemize}
\end{theorem}
Since Theorem \ref{thm:rates} follows immediately from Theorem \ref{thm:dimensionforrates}, the focus of this work will be to prove Theorem \ref{thm:dimensionforrates}. The proof of this theorem, and many related results, are presented in the remainder of this paper.

As a warm-up we consider the simplest case.
\begin{theorem}
\label{thm:linear}
    If $\VC(\H)=\infty$ then $\H$ is not learnable, i.e. $\Mopt(\H, n) \in \Omega(n)$.
\end{theorem}
\begin{proof}
Let $n \in \mathbb{N}$ and let $S$ be a shattered set of size $n$. The adversary then gives $S$ to the learner and chooses its labels uniformly at random. Since $S$ is shattered by $\H$, these labels are realizable. The learner then averages half a mistake per round, so $\Mopt(\H, n) \geq n/2 \in \Omega(n)$.
\end{proof}
\subsection{Upper Bounds}
\label{sec:upper}
We now consider the interesting regimes of $\VC(\H)<\infty$, $\Sdim(\H)<\infty$, or $\SDdim(\H)<\infty$. \footnote{It is straightforward to see that $\VC(\H)\le \Sdim(\H)$ due to their respective definitions. Because a shattered set of size $n$ can force $n$ mistakes for a deterministic self-directed learner, we also have $\VC(\H) \le \SDdim(\H)$.} We first show the upper bounds.
\subsubsection{\texorpdfstring{$\VC(\H)<\infty$}{VC finite}}
\begin{algorithm}[H]
\caption{\textsc{Prune-and-Eat}$(\H,S)$}
\label{alg:prune-and-eat}
\begin{algorithmic}[1]
\State $\mathcal V \gets \H|_S, n \gets |S|$ \Comment{version space: projection of $\H$ onto $S$}
\State \textbf{while} {$S$ is not empty}:
    \State \hspace{\algorithmicindent} $\mathcal J \gets \{h \in \mathcal V : \sum_{x \in S} h(x) \le \sqrt{\VC(\H)\, n \log n}\}$ \Comment{sparse concepts}
    \State \hspace{\algorithmicindent} $\mathcal V \gets \mathcal V \setminus \mathcal J$
    \State \hspace{\algorithmicindent} \textbf{if} $\V=\emptyset$  \textbf{then break}
    \State \hspace{\algorithmicindent} $a \gets \argmax_{x \in S} \sum_{h \in \mathcal V} h(x)$ \Comment{best apple}
    \State \hspace{\algorithmicindent} Predict $\hat y = 1$ on $a$ and receive the true label $y$
    \State \hspace{\algorithmicindent} $S \gets S \setminus \{a\}$
    \State \hspace{\algorithmicindent} $\mathcal V \gets \{h \in \mathcal V : h(a) = y\}|_S$
    \State \textbf{if } $|S|>0$ \textbf{ then} predict $\hat y=0$ on every remaining $x \in S$. \Comment{no further feedback}
\end{algorithmic}
\end{algorithm}
We first consider the regime in which $\VC(\H)<\infty$.
We design an algorithm that helps the learner select relatively safe points to query while progressively eliminating concepts that are either inconsistent with the true concept or which can be safely ignored.

At a high level, \textsc{Prune-and-Eat} alternates between two steps. 
First, it discards any concepts in the current version space that label too few of the remaining points as $1$.
This is safe because if the true concept labels almost all remaining points as $0$, the learner can afford to predict $0$ on all of them and still make few mistakes.
Second, among the surviving hypotheses, the algorithm greedily selects the point on which the largest fraction of the concepts predicts $1$, and queries its label by predicting it label to be $1$.
Pruning guarantees that this point is labeled $1$ by a $t$-fraction of the version space for $t=\sqrt{\VC(\H) \log n / n}$.
Hence, whenever eating it results in a mistake, that entire $t-$fraction of concepts is inconsistent with the revealed label and is eliminated.
We now state and prove this formally.
\begin{theorem}
\label{thm:vc-apple-tasting}
Let $\H$ be a concept class with $\VC(\H) < \infty$. Then $\H$ is learnable by algorithm \textnormal{\textsc{Prune-and-Eat}}.
Furthermore, we obtain a bound on the optimal number of mistakes 
\begin{equation*}
    M_{\mathrm{optimal}}(\H,n) \le 2\sqrt{\VC(\H)\, n \log n} + 1.
\end{equation*}
\end{theorem}

\begin{proof}
Let $S = (x_1, \ldots, x_n)$ be the points chosen by the adversary. 
Let $h^* \in \H$ denote the (unknown) true labeling concept.
Let $\V_0=\H|_S$ be the initial version space, and for $i\ge 1$ let $\V_i$ denote the version space at the start of round $i$ and $\V_i^\mathrm{pruned}$ its state immediately after the pruning step of that round.
Let $S_i$ denote the data set in the $i^\mathrm{th}$ round.
Every mistake the algorithm makes is of exactly one of two types, according to its prediction:
\begin{itemize}
    \item $M_1$: mistakes on points the algorithm eats, i.e. $\hat y=1$ while $h^*(a)=0$;
    \item $M_2$: mistakes on points the algorithm defaults to $0$ on after exiting the loop i.e. $\hat y=0$ while $h^*(x)=1$.
\end{itemize}
Since the algorithm predicts $1$ exactly on the points it eats inside the loop and $0$ on every point remaining at exit, these two classes partition the mistake rounds, so the total number of mistakes is $M_1 + M_2$.

\vspace{1em}

\noindent\textbf{\emph{Key observation}.}
Fix a round $i$ and consider the version space $\V_i$ before pruning.
We remove the concepts that are sparse: i.e., if the number of points in $S$ a concept labels as $1$ is less than $\sqrt{\VC(\H) n \log n}$, then we remove that concept to obtain $\V_i^\mathrm{pruned}$.
We note that by construction every $h \in \V^\mathrm{pruned}_i$ satisfies $\sum_{x \in S} h(x) > \sqrt{\VC(\H) n \log n}$. Summing over all $h \in \V^\mathrm{pruned}_i$ and exchanging the order of summation,
\begin{equation*}
    \sum_{x \in S_i} \sum_{h \in \V^\mathrm{pruned}_i} h(x) = \sum_{h \in \V^\mathrm{pruned}_i} \sum_{x \in S_i} h(x) > \lrabs{\V^\mathrm{pruned}_i} \cdot \sqrt{\VC(\H) n \log n}.
\end{equation*}
Since $|S_i| \le n$, averaging over the points of $S_i$ show that some $x \in S_i$ satisfies
\begin{equation*}
    \sum_{h \in \V^\mathrm{pruned}_i} h(x) > \frac{\lrabs{\V^\mathrm{pruned}_i}}{n} \cdot \sqrt{\VC(\H) n \log n} = \lrabs{\V^\mathrm{pruned}_i} \cdot \sqrt{\frac{\VC(\H) \log n}{n}}.
\end{equation*}
Therefore after pruning, some point in $S$ is labeled $1$ by at least a $t = \sqrt{\frac{\VC(\H) \log n }{n}}$ fraction of the surviving hypotheses. 
In particular, the point $a$ chosen by the algorithm, which maximizes this count, satisfies this bound as well.
Note that we can assume $t < 1$; if $t \geq 1$, then $n \leq \sqrt{\VC(\H) n \log n}$, so the claimed mistake bound is trivial.

\vspace{1em}

\noindent\textbf{Bounding $M_1$.} 
Suppose the algorithm eats $a$ in round $i$ and makes a mistake, so the revealed label is $y = 0$. By the key observation, more than a $t$-fraction of $\V^{\mathrm{pruned}}_i$ labels $a$ as $1$; every such concept is inconsistent with $y$ and is removed. 
Hence
\begin{equation*}
    |\V_{i+1}| \le (1-t)\lrabs{\V_i^{\mathrm{pruned}}} \le (1-t)|\V_i| ,
\end{equation*}
using $\V_i^{\mathrm{pruned}} \subseteq \V_i$. Version spaces are non-increasing across all other rounds, so we can chain this inequality over the $M_1$ rounds on which an eating mistake occurs. Let $i^*$ denote the last round completed before the loop terminates. The loop breaks only upon \emph{detecting} that $\V$ is empty, so the version space at the start of that round i.e. $\V_{i^*}$ is nonempty, giving $|\V_{i^*}| \ge 1$. Hence,
\begin{equation*}
    1 \le |\V_{i^*}|\le (1-t)^{M_1-1} |\V_0|.
\end{equation*} 
We now use Sauer's Lemma, giving $|\mathcal V_0| \le n^{\VC(\H)}$, and take logarithms to obtain
\begin{equation*}
    \VC(\H) \log n + (M_1-1) \log(1-t) \ge 0.
\end{equation*}
Using $\log(1-t) \le -t$ for $t \in (0,1)$,
\begin{equation*}
    \VC(\H) \log n \ge (M_1-1) t = (M_1-1) \sqrt{\frac{\VC(\H) \log n}{n}}
    \implies
    M_1 \le \sqrt{\VC(\H) n \log n} + 1.
\end{equation*}

\vspace{1em}

\noindent\textbf{Bounding $M_2$.} If the loop terminates with $S = \emptyset$ then every point was eaten, no point receives the default prediction, and $M_2 = 0$. So suppose the loop terminates with $\V = \emptyset$ in some round $j$. Since $h^*$ is consistent with every revealed label, it is stays in the version space; the only step that can remove it is pruning. Hence if $M_2 > 0$, then $h^* \in \mathcal J$ at some round $k \le j$, which
means $\sum_{x \in S_k} h^*(x) \le \sqrt{\VC(\H)n \log n}$.
The points receiving the default prediction $0$ are exactly those remaining at termination,
and $S_j \subseteq S_k$ since $S$ only shrinks. Each such mistake occurs at a point $x$ with
$h^*(x) = 1$, so
\begin{equation*}
    M_2 = \sum_{x \in S_j} h^*(x) \le \sum_{x \in S_k} h^*(x) \le \sqrt{\VC(\H)n \log n} .
\end{equation*} 

\noindent\textbf{Total mistakes.} Combining the two bounds,
\begin{equation*}
    M_{\mathrm{optimal}}(\H,n) \le M_1 + M_2 \le 2\sqrt{\VC(\H) n \log n} + 1.
\end{equation*}
\end{proof}
We now consider the case where $\Sdim(\H)<\infty$.
The following theorem will help us resolve the $O(\log n)$ and $O(1)$ regimes at the same time.
\begin{theorem}
\label{thm:upperboundlognconstant}
    For a concept class $\H$,
    \[\Mopt(\H,n) \le \min\lrset{\Sdim(\H)\cdot\SDdim(\H),\frac{1}{c}\Sdim(\H)\log n}\]
\end{theorem}

We will prove Theorem \ref{thm:upperboundlognconstant} in two cases: $\SDdim(\H)=\infty$ and $\SDdim(\H)<\infty$.

\subsubsection{\texorpdfstring{$\SDdim(\H)=\infty, \Sdim(\H)<\infty$}{SDDim infinite}}.

\begin{algorithm}[H]
\caption{\textsc{Union-and-Eat}$(\H, S)$}
\label{alg:union-and-eat}
\begin{algorithmic}[1]
\State $\V \gets \U(\H)|_S$ \Comment{Projection of the union closure $\U(\H)$ as in Def. \ref{def:union-closure} onto $S$}
\State \textbf{while }{there exists $x \in S$, $h \in \V$ with $h(x) = 1$:}
    \State \hspace{\algorithmicindent} $a \gets \argmax_{x \in S} \sum_{h \in \V} h(x)$ \Comment{Best apple}
    \State \hspace{\algorithmicindent} Predict $\hat y = 1$ on $a$ and receive the true label $y$
    \State \hspace{\algorithmicindent} $S \gets S \setminus \{a\}$
    \State \hspace{\algorithmicindent} $\V \gets \{h \in \V : h(a) = y\}|_S$
    \State \textbf{if} $|S|>0$ \textbf{then} predict $\hat y=0$ on every remaining $x \in S$
\end{algorithmic}
\end{algorithm}

In this case, rather than working with $\V$ directly, we pass to its \emph{union closure}
$\U(\V)$, obtained by taking all pointwise-OR combinations of subsets of $\V$.
Learning $\H$ is no harder than learning $\U(\H)$ since $\H \subseteq \U(\H)$, so any mistake bound for $\U(\H)$ upper bounds the mistake bound for $\H$. 
The advantage of passing to $\U(\H)$ is that it is union-closed by construction, and union-closed families enjoy a useful averaging property: some point must be labeled $1$ by a constant fraction of the family. 
Consequently, greedily querying this point guarantees constant multiplicative progress on every mistake, in contrast to Algorithm \ref{alg:prune-and-eat}, where progress was only guaranteed after a pruning step. 
We now prove a few lemmas which will be helpful to prove an upper bound based on $\Sdim(\H)$.
\begin{lemma}
\label{lemma:Sdim-union-invariant}
For any concept class $\H,\Sdim(\H) = \Sdim(\U(\H)).$
\end{lemma}
\begin{proof}
$(\ge)$ Let $s = \Sdim(\U(\H))$, and let $P = \{p_1, \ldots, p_s\}$ be $\emptyset$-centered star-shattered in $\U(\H)$. Let $g_0 \in \U(\H)$ be the center: i.e. $g_0(p_j)=0$ for all $j \in [s]$. Let $g_1, \ldots, g_s \in \U(\H)$ be the witnesses: i.e. for all $i \in [s]$, we have $g_i(p_i) = 1$ and $g_i(p_j) = 0$ for all $j \neq i$.

We claim that $\H$ also $\emptyset$-centered star-shatters $P$. For the center, write $g_0 = h_{0, 1} \lor \cdots \lor h_{0, m_0}$ for some $h_{0, 1}, \ldots, h_{0, m_0} \in \H$. Because $g_0(p_j) = 0$ for every $j \in [s]$, and a union of concepts is $0$ at a point only if every concept in the union is $0$ there, each $h_{0, k_i}$ satisfies $h_{0, k_i}(p_j) = 0$ for all $j \in [s]$.

We can similarly extract the witnesses from the other $g_i$. Fix $i \in [s]$ and write $g_i = h_{i,1} \lor \cdots \lor h_{i,m_i}$ for some $h_{i,1}, \ldots, h_{i,m_i} \in \H$.
Since $g_i(p_j) = 0$ for every $j \neq i$, each $h_{i,k}$ satisfies $h_{i,k}(p_j) = 0$ for all $j \neq i$. 
Since $g_i(p_i) = 1$, at least one $h_{i,k_i} \in \H$ satisfies $h_{i,k_i}(p_i) = 1$. 
Combining these two facts, $h_{i,k_i} \in \H$ agrees with $g_0$ everywhere on $P$ except at $p_i$, where it equals $1$.
Repeating this for every $i \in [s]$ produces concepts
$h_{1,k_1}, \ldots, h_{s,k_s} \in \H$ that, together with $h_{0, 1} \in \H$,
$\emptyset$-centered star-shatter $P$. Hence $\Sdim(\H) \ge s = \Sdim(\U(\H))$.

$(\le)$ Since $\U(\H)$ contains the singleton union of each concept in $\H$, we have $\H \subseteq \U(\H)$. Hence any set $\emptyset$-centered star-shattered by $\H$ is also shattered by $\U(\H)$, so $\Sdim(\U(\H)) \ge \Sdim(\H)$.
\end{proof}
\begin{lemma}
\label{lemma:vc-star-equiv}
If $\H$ is a union-closed concept class then $\VC(\H)=\Sdim(\H)$.
\end{lemma}

\begin{proof}
$(\ge)$ Let $s = \Sdim(\H)$, and let $P = \{p_1,\ldots,p_s\}$ be $\emptyset$-centered star-shattered by center $h_0$ and witnesses $h_1,\ldots,h_s \in \U$: $h_0$ labels every point of $P$ as $0$, and $h_i$ agrees with $h_0$ except that $h_i(p_i) = 1$.

We claim that $\H$ shatters $P$. The concept $h_0$ realizes the all-zero labeling of $P$. For any nonzero target labeling $y = (y_1,\ldots,y_s) \in \{0,1\}^s$ of $P$, let $\mathcal W = \{h_i : y_i = 1\} \subseteq \{h_1,\ldots,h_s\}$.
Let $h_{\mathcal W} = \bigvee_{h \in \mathcal W} h$. Because $\H$ is union-closed and $\mathcal{W}$ is nonempty, we have $h_{\mathcal{W}} \in \H$.
Since each $h_i$ labels only $p_i$ as $1$ among $P$, $h_{\mathcal W}(p_i) = 1$ if and only if $h_i \in \mathcal W$ which is if and only if $y_i = 1$.
Hence $\H$ realizes every labeling of $P$, so $\VC(\H) \ge s$.

$(\le)$ Let $v = \VC(\H)$, and let $S' = \{q_1,\ldots,q_v\}$ be shattered by $\H$.
By shattering, for each $i \in [v]$ there is a concept $g_i \in \H$ realizing the labeling that assigns $q_i \mapsto 1$ and $q_j \mapsto 0$ for $j \neq i$. There is also a concept $g_0 \in \H$ realizing the all-$0$ labeling. 
Then $S'$ is $\emptyset$-centered star-shattered by center $g_0$ and witnesses $g_1,\ldots,g_v$, so $\Sdim(\H) \ge v$. 
\end{proof}
\begin{proposition}
\label{prop:union-closed}
There is a universal constant $c > 0$ such that for every finite union-closed class $\mathcal H$ of concepts (other than the trivial class), some point $x$ satisfies 
\[\lrabs{\{h \in \mathcal H : h(x) = 1\}} \ge c \cdot |\mathcal H|.\]
\end{proposition}
We use $c$ as a placeholder for the best constant available. 
At the time of writing \citet{gilmer2022constantlowerboundunionclosed} establish $c=0.01$, \citet{chase2022approximateunionclosedconjecture,pebody2022extensionmethodgilmer,sawin2023improvedlowerboundunionclosed,alweiss2024improvedlowerboundfrankls} establish $c=\frac{3-\sqrt{5}}{2}\approx 0.38197$, and finally \citet{liu2023improvinglowerboundunionclosed} establish $c = 0.38271$. 
Please refer to \cite{bruhn2013journeyunionclosedsetsconjecture} for the history of union-closed set conjecture (also known as \textit{Frankl's conjecture}) which suggests $c=\frac12$.
The exact value is not relevant for our purposes as long as $c>0$.

\begin{theorem}
\label{thm:Sdim-apple-tasting}
Let $\H$ be a concept class with $\Sdim(\H) < \infty$. Then $\H$ is learnable by algorithm \textnormal{\textsc{Union-and-Eat}}. Furthermore, we obtain a bound on the optimal number of mistakes
\[
    M_{\mathrm{optimal}}(\H,n) \le \frac{1}{c}\,\Sdim(\H) \log n,
\]
where $c \in (0, 1/2]$ is the universal constant from the union-closed set conjecture
(Proposition \ref{prop:union-closed}).
\end{theorem}

\begin{proof}
For $i\ge 1$, let $\V_i$ denote the version space at the start of the $i$th iteration of the while loop.
Since we only ever remove concepts for being inconsistent with the true concept, there is always a concept consistent with the true concept in $\V_i$.
As we only exit the while loop once the remaining concepts label all remaining points as $0$, we make no mistakes when labeling the remaining points $0$ in step 7.

That leaves the false positive mistakes made when predicting $1$ in step 4. We use Proposition \ref{prop:union-closed} to bound these mistakes, so we need to make sure that the version spaces $\V_i$ are all union-closed. We show this by induction on $i$.
As a base case, the initial version space $\V_1$ is the projection of the union-closed concept class $\U(\H)$ onto $S$, so it is union-closed.
Now assume $\V_i$ is union-closed for some fixed $i$. The next version space $\V_{i+1}$ is constructed in two steps:
\begin{itemize}
    \item First, it is restricted to the concepts which map $a$ to the true label $y$. Let $h_1, h_2$ be two such concepts. Since $\V_i$ is union-closed, we have $h_1 \lor h_2 \in \V_i$. Moreover, since $(h_1 \lor h_2)(a) = h_1(a) \lor h_2(a) = y \lor y = y$, it remains in the version space.
    \item Second, it is projected onto $S \setminus \{a\}$. As in the base case, this projection preserves union-closure.
\end{itemize}
Therefore we can guarantee that the point $a$ selected in each round of \textsc{Union-and-Eat} satisfies
$|\{h \in \V_i : h(a) = 1\}| \ge c \cdot |\V_i|$, where $c$ is the constant in Proposition \ref{prop:union-closed}.
Whenever the algorithm makes a mistake at $a$ its true label is $0$, so every $h \in \V_i$ with $h(a) = 1$ is inconsistent and is discarded from the next round's version space. Let $M$ be the number of mistakes made in total. Since no $\V_i$ is empty, we have: 
\begin{equation*}
    (1 - c)^M \cdot |\V_1| \ge 1,
\end{equation*}
By Lemmas \ref{lemma:Sdim-union-invariant} and \ref{lemma:vc-star-equiv}, we have $\VC(\V_1) = \Sdim(\V_1)= \Sdim(\H|_S) \le \Sdim(\H)$, so by Sauer's Lemma:
\begin{equation*}
    |\V_1| \le n^{\VC(\V_1)} \le n^{\Sdim(\H)}.
\end{equation*}
Taking logarithms and using $\log(1-c) \le -c$, we obtain
\begin{equation*}
    0 \le \Sdim(\H) \log n - c M
    \implies
    M \le \frac{1}{c} \Sdim(\H) \log n. 
\end{equation*}
So algorithm \textsc{Union-and-Eat} makes at most $\frac{1}{c} \Sdim(\H) \log n$ mistakes as desired.
\end{proof}
\subsubsection{\texorpdfstring{$\SDdim(\H)<\infty, \Sdim(\H)<\infty$}{SDDim finite}}
\begin{algorithm}[H]
    \caption{\textsc{First-Mistake}$(\V,(x_{1}, x_{2},\ldots,x_{b}),x_{b+1})$}
    \label{alg:firstmistake}
    \begin{algorithmic}[1]
        \State $\mathrm{mis}\gets \infty$
        \State \textbf{if} $x_{b+1} \neq \bot$ \textbf{then}
        \State \hspace{\algorithmicindent} Predict $\hat y=1$ for $x_{b+1}$ and obtain the true label $y$.
        \State \hspace{\algorithmicindent} \textbf{if} $y=1$ \textbf{then return} $\bot$.
        \State \hspace{\algorithmicindent} $\V \gets \V_{x_{b+1}\to 0}$, $\mathrm{mis}\gets b+1$
        \State Let $f(h) = \min\lrset{j \in [b]: h(x_{j})=1}$ for all $h \in \V$, with $\min \emptyset = \infty$.
        \State \textbf{while} there is some $h \in \V$ such that $f(h)<\mathrm{mis}$ \textbf{do}
        \State \hspace{\algorithmicindent} Let $j = \max\lrset{f(h):h \in \V, f(h) <\mathrm{mis}}$
        \State \hspace{\algorithmicindent} Predict $\hat y=1$ at $x_{j}$ and obtain the true label $y$.
        \State \hspace{\algorithmicindent} $\V \gets \V_{x_{j}\to y}$
        \State \hspace{\algorithmicindent} \textbf{if} $y=1$ \textbf{then} $\mathrm{mis}\gets j$
        \State \textbf{if} $\mathrm{mis}=\infty$ \textbf{then return} $\bot$ \textbf{else return} $x_{\mathrm{mis}}$
    \end{algorithmic}
\end{algorithm}

\begin{algorithm}[H]
    \caption{\textsc{Simulate-and-Eat}$(\H,S,\sdsolver)$}
    \label{alg:simulate-and-eat}
    \begin{algorithmic}[1]
        \State $\V \gets \H|_S$
        \State \textbf{while} $\V$ contains concepts that disagree at a point in $S$ \textbf{do}:
        \State \hspace{\algorithmicindent} Let $n \gets |S|,S^0 \gets \emptyset, x_{\mathrm{mis}} \gets \bot$
        \State \hspace{\algorithmicindent} Simulate a run of $\sdsolver(\V,S)$ assuming all predictions are correct.
        \State \hspace{\algorithmicindent} Let the order and the predictions in the previous step be $\sigma$,
        $\lr{\hat{y}_{\sigma(1)},\ldots, \hat{y}_{\sigma(n)}}$ resp.
        \State \hspace{\algorithmicindent} Let $x_{\sigma(n+1)}=\bot,\hat{y}_{\sigma(n+1)}=1$.
        \State \hspace{\algorithmicindent} \textbf{for} $i$ goes from $1$ to $n+1$ \textbf{do}
        \State \hspace{\algorithmicindent} \hspace{\algorithmicindent} \textbf{if} $\hat{y}_{\sigma(i)}=0$ \textbf{then} $S^0 \gets S^0 \circ\lrset{x_{\sigma(i)}}$
        \State \hspace{\algorithmicindent} \hspace{\algorithmicindent} \textbf{if} $\hat{y}_{\sigma(i)}=1$ \textbf{then}
        \State \hspace{\algorithmicindent}\hspace{\algorithmicindent}\hspace{\algorithmicindent} $x_{\mathrm{mis}} \gets$ \textsc{First-Mistake}$\lr{\V,S^0,x_{\sigma(i)}}$
        \State \hspace{\algorithmicindent}\hspace{\algorithmicindent}\hspace{\algorithmicindent} \textbf{if} $x_{\mathrm{mis}}\neq \bot$ \textbf{then}
        \State \hspace{\algorithmicindent}\hspace{\algorithmicindent}\hspace{\algorithmicindent} \hspace{\algorithmicindent} Let $R$ be the set of points before $x_{\mathrm{mis}}$ in $S$ w.r.t. order $\sigma$.
        \State \hspace{\algorithmicindent}\hspace{\algorithmicindent}\hspace{\algorithmicindent} \hspace{\algorithmicindent} Label all the points in $R$ according to the labels given by $\sdsolver$.
        \State \hspace{\algorithmicindent}\hspace{\algorithmicindent}\hspace{\algorithmicindent} \hspace{\algorithmicindent} $S \gets S \setminus (R\cup \lrset{x_{\mathrm{mis}}})$
        \State \hspace{\algorithmicindent}\hspace{\algorithmicindent}\hspace{\algorithmicindent} \hspace{\algorithmicindent} $\V \gets \lrset{h \in \V:h(x_i)=\hat{y}_i \text{ for all } x_i \in R \text{ and } h(x_{\mathrm{mis}})=1-\hat y_{\mathrm{mis}}}$
        \State \hspace{\algorithmicindent}\hspace{\algorithmicindent}\hspace{\algorithmicindent} \hspace{\algorithmicindent} \textbf{break}
        \State \hspace{\algorithmicindent} \textbf{if} $x_{\mathrm{mis}} = \bot$ \textbf{then} 
        \State \hspace{\algorithmicindent} \hspace{\algorithmicindent} Label all the points in $S$ according to the labels given by $\sdsolver$ and \textbf{break}
        \State Predict all remaining points according to the concepts in $\V$
        \end{algorithmic}
\end{algorithm}

In this case, the idea is a kind of reduction from apple-tasting feedback to the full-information case.
We make $\Sdim(\H)$ mistakes to recognize the first position at which the full-information learner makes a mistake.
We know that in full-information, we can obtain the true concept in $\SDdim(\H)$ mistakes. Therefore, we obtain the true concept in at most $\Sdim(\H) \times \SDdim(\H)$ mistakes with apple-tasting feedback.

The following lemma shows that we can obtain the first position where the full-information (self-directed) learner makes a mistake using apple-tasting feedback with at most $\Sdim(\H)$ mistakes.
\begin{lemma}
\label{lemma:firstmistake}
Let $(x_1,\ldots,x_b), x_{b+1}$ be points provided to \textnormal{\textsc{First-Mistake}}.
Let $\V$ be a version space such that $h^* \in \V$ and there exists $h^b \in \V$ such that $h^b(x_i)=0$ for all $ i \in [b]$ and, if $x_{b+1} \neq \bot$, then $h^b(x_{b+1})=1$. 
Let \[j^*=\min\lrset{j \in [b]:h^*(x_j)=1} \text{ or } \infty \text{ if no such point exists}\] 
Then one of these three cases occurs:
\begin{itemize}
    \item If $h^*(x_{b+1})=1$ then \textnormal{\textsc{First-Mistake}} returns $\bot$ and makes no mistakes.
    \item If $h^*(x_{b+1})=0$ then \textnormal{\textsc{First-Mistake}} returns $x_{\min\lrset{j^*,b+1}}$ and makes at most $\Sdim(\V)$ mistakes.
    \item If $x_{b+1}=\bot$ then \textnormal{\textsc{First-Mistake}} returns $x_{j^*}$ if $j^*<\infty$ and $\bot$ otherwise. It also makes at most $\Sdim(\V)$ mistakes.
\end{itemize}
\end{lemma}

\begin{proof}
Throughout, every update is of the form $\V \gets \V_{x \to y}$ where $y$ is the true label revealed at $x$; hence $h^* \in \V$ at all times, and $\V$ only shrinks. 
Note also that $f(h) = \min\lrset{j \in [b] : h(x_j) = 1}$ depends only on $h$, so it is unaffected by these updates.

\emph{Case $1$: $h^*(x_{b+1}) = 1$.}
In this case, the lemma statement is immediate from the algorithm: line 3 predicts $\hat y = 1$ at $x_{b+1}$ and the revealed label is $h^*(x_{b+1}) = 1$, so line 4 returns $\bot$ immediately and we make no mistakes.

In the other two cases, the algorithm does non-trivial work. We first make two observations about this work by linking the variable $\mathrm{mis}$ to $j^*$:
\begin{enumerate}
    \item $\mathrm{mis} \ge j^*$ when $j^*<\infty$: $\mathrm{mis}$ is lowered only at line 11, upon a probe at $x_j$ revealing $y = 1$, and then $h^*(x_j) = 1$ forces $j \ge j^*$. 
    In case $2$ the initial value $\mathrm{mis} = b+1$ satisfies this since $j^* \in [b]$ and in case $3$ the initial value is $\infty$.
    \item The loop cannot terminate while $\mathrm{mis} > j^*$: if $j^* < \infty$ then $h^* \in \V$ satisfies $f(h^*) = j^* < \mathrm{mis}$, so condition for the loop at line 7 holds.
\end{enumerate}
Combining the two, the loop terminates with $\mathrm{mis} = j^*$ whenever $j^* < \infty$.

\textbf{Correctness in cases $2$ and $3$}.

\emph{Case $2$: $h^*(x_{b+1}) = 0$.}
Line 3 incurs one mistake, line 5 sets $\mathrm{mis} \gets b+1$, and the loop runs. 
If $j^* \le b$ the observations give $\mathrm{mis} = j^*$ on termination, and line 12 returns $x_{j^*}$. If $j^* = \infty$, no probe can reveal $y=1$ (that would force $j \ge j^* = \infty$), so $\mathrm{mis}$ remains $b+1$ and line 12 returns $x_{b+1}$. 
Therefore, the returned point is $x_{\min\lrset{j^*,\,b+1}}$.

\emph{Case $3$: $x_{b+1} = \bot$.}
Here $\mathrm{mis} = \infty$ on entry to the loop. If $j^* < \infty$, then $\mathrm{mis} = j^*$ on termination and line 12 returns $x_{j^*}$. If $j^* = \infty$, then no probe reveals $y = 1$, so $\mathrm{mis}$ remains $\infty$ and line 12 returns $\bot$.

\textbf{Mistake bound for cases $2$ and $3$.}

A probe is a mistake exactly when the revealed label is $0$. Let $j_1 > j_2 > \cdots > j_k$ be the indices in $[b]$ of the loop probes revealing $0$, and in Case $2$ put additionally $j_0 := b+1$, the index of the failed initial probe. 
Set $P = \lrset{x_{j_i}}$, over $i \in \lrset{0,1,\ldots,k}$ in Case 2 and $i \in \lrset{1,\ldots,k}$ in Case $3$. 
The number of mistakes is exactly $|P|$, so it suffices to show that the initial $\V$ $\emptyset$-centered star-shatters $P$.

\emph{The center.} Each $j_i$ was a probe revealing the true label $0$, so $h^*(x_{j_i}) = 0$ for every $i$, and $h^* \in \V$ by hypothesis. Hence $h^*$ labels every point of $P$ as $0$. 

\emph{The witnesses.} For $i = 0$ (case $2$ only) take $h^b$, which by hypothesis lies in $\V$, satisfies $h^b(x_{b+1}) = 1$, and satisfies $h^b(x_j) = 0$ for every $j \in [b]$, hence in particular at $x_{j_1},\ldots,x_{j_k}$.

For $i \ge 1$, let $\V_i \subseteq \V$ denote the version space immediately before the $i$-th failed loop probe. 
By the selection rule at line 8 there is $h_i \in \V_i$ with $f(h_i) = j_i$. 
Then
\begin{itemize}
    \item $h_i(x_{j_i}) = 1$, directly from $f(h_i) = j_i$.
    \item $h_i(x_{j_{i'}}) = 0$ for every $i' > i$, since $j_{i'} < j_i = f(h_i)$ and $f(h_i)$ is the least index of $[b]$ at which $h_i$ takes the value $1$.
    \item $h_i(x_{j_{i'}}) = 0$ for every $i' < i$, including $i' = 0$ in case $2$, since $h_i \in \V_i$ and each of those earlier probes revealed the label $0$, after which line 10 (respectively line 5) discarded every concept taking the value $1$ there.
\end{itemize}
So $h_i$ is $1$ at $x_{j_i}$ and $0$ on $P \setminus \lrset{x_{j_i}}$, and $h_i \in \V$
because $\V_i \subseteq \V$.

Together with the center $h^*$, these witnesses $\emptyset$-centered star-shatter $P$, hence $|P| \le \Sdim(\V)$ and the algorithm makes at most $\Sdim(\V)$ mistakes.
\end{proof}

\begin{theorem}
\label{thm:sddimupperbound}
    Let $\H$ be a concept class with $\Sdim(\H)<\infty$ and $\SDdim(\H)<\infty$. Then $\H$ is learnable by algorithm \textnormal{\textsc{Simulate-and-Eat}}. Furthermore, we also obtain a bound on the optimal number of mistakes
    \[\Mopt(\H,n) \le \Sdim(\H) \cdot \SDdim(\H)\]
\end{theorem}
\begin{proof}
    We use induction on $\SDdim(\H)$. As a base case, if $\SDdim(\H) = 0$, then $\H$ contains only one concept, so the main loop terminates immediately and predicts all points correctly.

    Now assume that, for some $d \geq 0$, \textsc{Simulate-and-Eat} is correct and achieves the desired mistake bound on all concept classes $\H$ with $\SDdim(\H) \leq d$. Let $\H'$ be an arbitrary concept class with $\SDdim(\H') = d+1$, let $S$ be a multiset, and let $h^* \in \H'$ be the true labeling concept.
    
    We now run \textsc{Simulate-and-Eat} on $\H'$ and $S$. Let $\sigma$ be the order in which $\sdsolver$ probes the points in $S$ and let $\lr{\lr{x_{\sigma(1)},\hat y_{\sigma(1)}}}, \ldots, \lr{\lr{x_{\sigma(n)}, \hat y_{\sigma(n)}}}$ denote the labels provided by $\sdsolver$ assuming that all the predictions it made are correct. We will use Lemma \ref{lemma:firstmistake} to characterize the behavior of the calls to \textsc{First-Mistake}, so we need to show that its conditions hold:
    \begin{itemize}
        \item We always have $h^* \in \V$ since $h^* \in \H'$.
        \item Since $\sdsolver$ is consistent, the concept $h$ it follows satisfies $h(x)=0$ for all $x \in S^0$ and $h(x_{\sigma(i)})=1$ for all $i$ such that $\hat{y}_{\sigma(i)} = 1$. This concept satisfies the condition of $h^b$ in Lemma \ref{lemma:firstmistake}.
    \end{itemize}

    We now briefly handle the case where all labels provided by $\sdsolver$ are correct. Here, the only call to \textsc{First-Mistake} which can make a mistake is the final one with $x_{\sigma(n+1)} = \bot$. By case 3 of Lemma \ref{lemma:firstmistake}, this call will make at most $\Sdim(\H')$ mistakes and return $\bot$. Line 17 then correctly labels all points as $\sdsolver$ did. Since $d+1 \geq 1$, we make at most $\Sdim(\H') \cdot \SDdim(\H')$ mistakes as desired.

    Now that at least one $\hat{y}_{\sigma(i)}$ is incorrect, let $i^*$ be the first index for which $h^*(x_{\sigma(i^*)}) \neq \hat{y}_{\sigma(i^*)}$. We claim that $x_{\mathrm{mis}}$ will be set to $x_{\sigma(i^*)}$, so that the updates on lines 14 and 15 correctly assign labels to $x_{\sigma(1)}, \ldots, x_{\sigma(i^*)}$. Let $i$ be the first index for which the call $\textsc{First-Mistake}(\V, S^0, x_{\sigma(i)})$ does not fall into case $1$ of Lemma \ref{lemma:firstmistake}; such an index exists because $x_{\sigma(n+1)} = \bot$. We have $i \geq i^*$ because if $i \leq n$, then $h^*(x_{\sigma(i)}) = 0 \neq \hat{y}_{\sigma(i)}$. We then have two cases based on $\hat{y}_{\sigma(i^*)}$:
    \begin{itemize}
        \item If $\hat{y}_{\sigma(i^*)} = 1$, then we have $i \leq i^*$, so $i=i^*$. The call to \textsc{First-Mistake} therefore falls under case 2 of Lemma \ref{lemma:firstmistake}. At this time, the true labels of all points in $S^0$ are $0$, so \textsc{First-Mistake} returns $x_{\sigma(i^*)}$ as desired and makes at most $\Sdim(\H')$ mistakes.
        \item If $\hat{y}_{\sigma(i^*)} = 0$, then $x_{\sigma(i^*)}$ is the first incorrectly labeled point in $S^0$. We may be in either case 2 or 3 of Lemma \ref{lemma:firstmistake}, but in either case this guarantees that $x_{\sigma(i^*)}$ is returned and at most $\Sdim(\H')$ mistakes are made.
    \end{itemize}

    Therefore, lines 14 and 15 correctly label the prefix $x_{\sigma(1)}, \ldots x_{\sigma(i^*)}$ and update the version space accordingly. Let $\H$ be this new version space. By the construction of $\sdsolver$, we have $\SDdim(\H) \leq \SDdim(\H' - 1) = d$. The next iteration of the loop is indistinguishable from running \textsc{Simulate-and-Eat} with $\H$ as the starting concept class, so by the inductive hypothesis it makes at most $\Sdim(\H) \cdot \SDdim(\H) \leq \Sdim(\H') \cdot d$ more mistakes. Adding in the $\Sdim(\H')$ mistakes from the first iteration we get at most $\Sdim(\H') \cdot \SDdim(\H')$ mistakes, completing the proof.
\end{proof}
\begin{proof}[Proof of Theorem \ref{thm:upperboundlognconstant}]
    We can combine Theorem \ref{thm:Sdim-apple-tasting} and Theorem \ref{thm:sddimupperbound} to obtain $\Mopt(\H,n) \le \min\lrset{\Sdim(\H)\SDdim(\H),\frac1c \Sdim(\H)\log n}$.
\end{proof}
\subsection{Lower Bounds}
\label{sec:lower}
We now take a look at the lower bounds.
Most of our lower bounds are based on an extension of the singletons concept class which we call \emph{block singletons}. 
Suppose a concept class $\H$ satisfies $\Sdim(\H) \geq d$ for some $d \in \mathbb{N}$. 
We then make a multiset of instances $S$ by including $d$ copies of each of the $d$ elements of an $\emptyset$-centered star-shattered set. 
This set is given by the adversary to the learner at the beginning of the game.
By definition, for each of these blocks of $d$ apples there exists a concept in $\H$ that labels only that block as $1$. 
Intuitively, any self-directed learner with apple-tasting feedback should make roughly $d$ mistakes on this multiset of instances: either they eat apples from few blocks (and miss the $d$ that are labeled $1$), or they eat apples from many blocks (including about $d$ that are labeled $0$).

\begin{theorem}
\label{lemma:sddimlowerbound}
    Let $d \in \mathbb{N}$ and let $\H$ be a concept class. If $\Sdim(\H) \geq d$, then there exists a multiset $S$ of size $d^2$ on which any randomized self-directed learner with apple-tasting feedback makes at least $\frac{2d^2}{3d+1}$ mistakes in expectation.
\end{theorem}

\begin{proof}
    Let $\{x_1, \ldots, x_d\}$ be a set that $\H$ $\emptyset$-centered star-shatters. 
    Let $S$ be a multiset consisting of $d$ copies of each of $x_1, \ldots, x_d$. 
    This multiset is given by the adversary to the learner at the beginning of the game.
    By definition, there exist concepts $h_1, \ldots, h_d \in \H$ such that, for each $i \in \{1, \ldots, d\}$, we have $h_i(x_i) = 1$ and $h_i(x_j) = 0$ for $j \neq i$. 
    Each of these concepts has one \emph{good} block labeled $1$ and $d-1$ \emph{bad} blocks labeled $0$. 
    There also exists a concept $h_0$ for which every block is bad. 
    We have the adversary choose the following distribution over these concepts at the beginning of the game:
    \begin{itemize}
        \item With probability $\frac{2d}{3d+1}$, choose one of $h_1, \ldots, h_d$ uniformly at random.
        \item With probability $\frac{d+1}{3d+1}$, choose $h_0$.
    \end{itemize}

    We claim that under this distribution no randomized learner can make fewer than $\frac{2d^2}{3d+1}$ mistakes on average. 
    
    Let $\mathcal{A}$ be an arbitrary randomized learning algorithm. We make a few observations which will aid our analysis.
    \begin{itemize}
        \item We can assume without loss of generality that $\mathcal{A}$ samples one point from each of the blocks in an arbitrary order and predicts $1$ on them until it either finds a good block or decides to stop; any other strategy makes unnecessary mistakes.
        \item If the algorithm stops because it found a good block, then it will predict $1$ for all the apples in that block and then predict $0$ for all other apples. This corresponds to the learner's discovery that the true concept is in $\{h_1,\ldots,h_d\}$.
        \item If the algorithm stops because it decided to terminate, then it will predict $0$ for all other apples. This corresponds to the learner's belief that the true concept is $h_0$.
        \item Since the only feedback $\mathcal{A}$ gets before stopping is a series of bad blocks, we can further assume that before making any samples it chooses some (possibly random) number $k$ of bad blocks to sample before stopping.
    \end{itemize} 
    We now compute the expected number of mistakes $\mathcal{A}$ makes for each fixed choice of $k \in \{0, \ldots, d\}$ which denotes the number of bad blocks the learner sampled from. 
    There are three possibilities, whose probabilities and mistake counts depend on $k$:
    \begin{itemize}
        \item In the first case the algorithm finds a good block at some $i\le k$. This indicates the true concept is in $\{h_1,\ldots,h_d\}$. In this case, the algorithm makes $i-1$ mistakes.
        \item In the second case the algorithm does not find a good block but the adversary had    chosen the true concept from $\{h_1,\ldots,h_d\}$. In this case, the algorithm makes $d+k$ mistakes.
        \item In the third case the algorithm does not find a good block and the adversary chose $h_0$ as the true concept. In this case, the algorithm makes $k$ mistakes.
    \end{itemize}
    We now combine these three cases to compute the expected number of mistakes made by the algorithm.
    \begin{align*}
        \E [\text{$\mathcal{A}$'s mistakes} \mid k] &= \underbrace{\frac{2d}{3d+1}\cdot \frac{1}{d} \cdot \sum_{i=1}^k (i-1)}_{\text{first case}} + \underbrace{\frac{2d}{3d+1}\cdot \frac{d-k}{d}\cdot(d+k)}_{\text{second case}}+\underbrace{\frac{d+1}{3d+1}\cdot k}_{\text{third case}} \\
        &=  \frac{2d^2}{3d+1} + \frac{k(d-k)}{3d+1}
    \end{align*}
    Since $0 \le k\le d, k(d-k)\ge 0$, so we have:
    \begin{align*}
        \E [\text{$\mathcal{A}$'s mistakes} \mid k] &\geq \frac{2d^2}{3d+1}
    \end{align*}
    Since this inequality holds for all $k$, it holds for any distribution over $k$, and therefore $\mathcal{A}$ makes at least $\frac{2d^2}{3d+1}$ mistakes in expectation.
\end{proof}
Note that the adversary in Lemma \ref{lemma:sddimlowerbound} commits to a distribution over the true concepts before beginning the game with the learner. 
This adversary is weaker than the general case, since our model allows choosing the labels adaptively.
We call this an \emph{oblivious} adversary.
We also refer readers to section \ref{sec:obliviousadversary} where we obtain a tight upper bound for a game against an oblivious adversary when $\VC(\H)<\infty$.

Since this block singleton structure exists in any concept class with $\Sdim(\H)\ge d$, it is especially powerful when $\Sdim(\H) = \infty$. We use this idea to show the following theorem.
\begin{theorem}
\label{thm:sqrtnlowerbound}
    For any concept class $\H$, if $\Sdim(\H)= \infty$ then $\Mopt(\H,n) \in \Omega(\sqrt{n})$.
\end{theorem}
\begin{proof}
    Let $\H$ be a concept class with $\Sdim(\H)= \infty$. 
    For any fixed $n \in \mathbb{N}$, let $m = \lfloor \sqrt{n} \rfloor$. Then $\Sdim(\H)\ge m$, so by Lemma \ref{lemma:sddimlowerbound}  there exists a sample set of size at most $m^2 \leq n$ on which any randomized self-directed learner with apple-tasting feedback makes at least $\frac{2m^2}{3m+1} \in \Omega(\sqrt{n})$ mistakes.
\end{proof}

When $\Sdim(\H) < \infty$, Lemma \ref{lemma:sddimlowerbound} yields only a constant lower bound. We supplement it with a bound based on $\SDdim(\H)$. Because apple-tasting feedback is less useful than full-information feedback, the only difficulty is in showing that $\Omega(\SDdim(\H))$ also lower-bounds the performance of randomized algorithms. A full proof would involve retreading much of the ground of \cite{devulapalli2024dimensionselfdirectedlearning}, so we refer the reader to that paper for a more complete description of the trees involved.

\begin{theorem}
\label{thm:easylowerbound}
    Let $\H$ be a concept class such that $\Sdim(\H)$ and $\SDdim(\H)$ are both finite. Then for all sufficiently large $n$ we have $\Mopt(\H,n)\ge \frac{1}{2}\max\lrset{\Sdim(\H),\SDdim(\H)}$. Moreover, if $\SDdim(\H) = \infty$, then $\Mopt(\H, n) \in \omega(1)$.
\end{theorem}
\textbf{Informal proof}
    Apple-tasting feedback is strictly weaker than full-information: a full-information learner can simulate any apple-tasting learner by discarding the labels it would not have observed.
    Any lower bound for the full-information setting therefore transfers to the apple-tasting setting, so we consider the performance of a full-information learner.
    \citet{devulapalli2024dimensionselfdirectedlearning} characterize self-directed learning through the dimension $\SDdim(\H)$, which they show equals the optimal deterministic mistake bound $M_{sd}(\H)$. 
    Combinatorially, $\SDdim(\H)$ can be constructed as the largest $k$ for which there is a set $S$ and a realizable \emph{$k$-tree} on $(\H,S)$: a tree whose nodes are subsets $V \subseteq S$, in which every $x \in V$ carries two outgoing edges labeled with distinct labels $y_x \neq y_x'$, every root-to-leaf path has length at least $k$, and every branch is realized by some concept of $\H$.

    In an analogous way to how  \citep{filmus2025optimalpredictionusingexpert} show bounds on randomized adversarial online learning using the Littlestone tree, we use a maximum $k$-tree to bound randomized self-directed learning.
    Here the adversary observes the learner's choice of point and distribution over predicted labels, but only sees the realized prediction after committing to the true label.

    If $\SDdim(\H) < \infty$, fix a realizable $k$-tree for $k=\SDdim(\H)$ and let the adversary play down it. 
    Whenever the learner selects a point $x$ of the current node $V$, two labels $y_x \neq y_x'$ are available, and at least one of them receives probability at most $1/2$ under the learner's distribution; the adversary answers with that label, forcing a mistake with probability at least $1/2$, and the play
    descends one level. 
    Since the learner must eventually predict on every point of $S$, and the points outside the current node are already determined by the tree, at least $k$ such descents occur. Hence any randomized learner suffers at least $k/2$ mistakes in expectation, giving $\Mopt \ge \frac12\SDdim(\H)$. If $\SDdim(\H) = \infty$ then we can take $k$ arbitrarily large for sufficiently large $n$, yielding the $\omega(1)$ bound.
    
    For the remainder of the lemma statement, Lemma \ref{lemma:sddimlowerbound} gives a lower bound of $\frac{2\Sdim(\H)^2}{3\Sdim(\H)+1} \geq \frac{1}{2}\Sdim(\H)$. Combing these two results give us the desired bound.
    \hfill
    $\blacksquare$

\subsection{Concluding the tetrachotomy result}
We can now prove our previously stated theorem on dimension to characterize the rates.
\begin{proof}[Proof of Theorem \ref{thm:dimensionforrates}]
Let a concept class $\H$ be given. We note that,
    \begin{itemize}
        \item By Theorem \ref{thm:linear}, if $\VC(\H)= \infty$ then $\Mopt(\H,n) \in \Omega(n)$.
        \item By Theorems \ref{thm:vc-apple-tasting} and \ref{thm:sqrtnlowerbound}, if $\VC(\H) < \infty$ and $\Sdim(\H)=\infty$ then $\Mopt(\H,n) \in \widetilde{\Theta}_{\H}(\sqrt{n})$.
        \item By Theorems \ref{thm:upperboundlognconstant} and \ref{thm:easylowerbound}, if $\Sdim(\H)<\infty$ and $\SDdim(\H)=\infty$ then $\Mopt(\H,n) \in O_{\H}(\log n) \cap \omega_{\H}(1)$.
        \item By Theorem \ref{thm:sddimupperbound}, if $\Sdim(\H)<\infty$ and $\SDdim(\H)<\infty$ then $\Mopt(\H,n) \in \Theta_{\H}(1)$.
    \end{itemize}
\end{proof}
\begin{proof}[Proof of Theorem \ref{thm:rates}]
Let a concept class $\H$ be given. By Definitions \ref{def:vc}, \ref{def:Sdim}, and \ref{def:SDdim} we have
$\VC(\H) \le \Sdim(\H)$ and $\VC(\H) \leq \SDdim(\H)$, so exactly one of the following holds:
$\VC(\H) = \infty$; or $\VC(\H) < \infty$ and $\Sdim(\H) = \infty$; or $\Sdim(\H)<\infty$ and $\SDdim(\H)=\infty$; or $\Sdim(\H) < \infty$ and $\SDdim(\H)<\infty$.
Theorem \ref{thm:dimensionforrates} assigns to each of these cases the rates
$\Omega(n)$, $\widetilde{\Theta}_{\H}(\sqrt n)$, $O_\H(\log n) \cap \omega_\H(1)$ and $\Theta_{\H}(1)$ respectively. 
Hence every concept class attains one of the four rates, and no class attains two, since the cases are mutually exclusive.
\end{proof}
\begin{remark}
    Self-directed online learning with apple-tasting feedback can be shown to allow learnability of more concept classes than its adversarial counterpart.
    For instance, the combinatorial characterization given by \cite{raman2024appletastingcombinatorialdimensions} shows that $\Omega(n)$ mistakes can be forced for thresholds in the adversarial online setting, but the $\VC-$dimension and $\emptyset$-centered star number $\Sdim$ are finite for this class. 
\end{remark}
\begin{remark}
    We note that $\SDdim(\H)$ and $\Sdim(\H)$ could be finite or infinite, independent of each other. Singletons on $\R$ have infinite $\Sdim(\H)$ but finite $\SDdim(\H)$. In section \ref{sec:classingap} we show that there also exists a class with finite $\Sdim(\H)$ but infinite $\SDdim(\H)$.
\end{remark}
\section{Agnostic Learning}
\label{sec:agnostic}
In the agnostic case, we obtain the same bounds that \cite{devulapalli2024dimensionselfdirectedlearning} did for full-feedback self-directed learning. Our lower bound is implied by theirs. Our upper bound uses the same strategy, but with the implementation of online learning with expert advice from \cite{raman2024appletastingcombinatorialdimensions} that works with apple-tasting feedback. We state both of these prior results here.

\begin{proposition}[Theorem 17 from \cite{devulapalli2024dimensionselfdirectedlearning}]
\label{prop:agnosticlower}
    Let $\H$ be a concept class with finite VC dimension. Then for all $n \in \mathbb{N}$, there exists an instance space of size $n$ on which any self-directed learning algorithm, with full-information feedback, has expected regret $\Omega(\sqrt{n \VC(\H)})$. \label{prop:devulapalli-self-directed-agnostic-lower-bound}
\end{proposition}

\begin{proposition}[Theorem 16 from \cite{raman2024appletastingcombinatorialdimensions}]
    \label{prop:raman-online-agnostic-apple-tasting}
    Let $N, T \in \mathbb{N}$ and let $\mathcal{E}^1, \ldots, \mathcal{E}^{N}$ be experts. For expert $\mathcal{E}^i$, let $\mathcal{E}^i_1, \ldots, \mathcal{E}^i_{T}$ be their predictions in rounds $1, \ldots, T$. Let $y_1, \ldots, y_T$ be the sequence of true labels. Then there exists an online learning algorithm that receives only apple-tasting feedback and outputs predictions $\hat{y}_1, \ldots, \hat{y}_T$ that satisfy:
    \begin{align*}
        \E \left[ \sum_{t=1}^T \1 \{\hat{y}_t \neq y_t\} \right] &\leq \left(\inf_{j \in [N]} \sum_{t=1}^T \1 \{\mathcal{E}^j_t \neq y_t\}\right) + \sqrt{3T \ln N}
    \end{align*}
\end{proposition}

\begin{theorem}
\label{thm:agnostic}
    Let $\H$ be a concept class with finite VC dimension. The optimal expected regret of learning $\H$ in the self-directed, apple-tasting, agnostic setting satisfies:
    \begin{align*}
        \Ropt(\H,n) \in \Omega(\sqrt{n \VC(\H)}) \qquad \text{and} \qquad \Ropt(\H,n) \in O(\sqrt{n \VC(\H) \log n})
    \end{align*}
\end{theorem}

\begin{proof}
    The lower bound is implied by Proposition \ref{prop:devulapalli-self-directed-agnostic-lower-bound}, since apple-tasting feedback is weaker than full-information feedback.

    For the upper bound, we use the algorithm of Proposition \ref{prop:raman-online-agnostic-apple-tasting}. After receiving the initial set of points $S$, the learner instantiates one expert for each projection of the concept class onto $S$. The performance of the best expert is then equal to the performance of the best concept, and by Sauer's lemma we have $\log N \in O(\VC(\H) \log n)$. The learner chooses an arbitrary order of the points in $S$ and commits to predicting on them in that order. The regret is therefore $O(\sqrt{n \log N}) = O(\sqrt{n \VC(\H) \log n})$.
\end{proof}
\section{Oblivious adversaries}
\label{sec:obliviousadversary}
Throughout our paper, we have been considering the most powerful adversary possible in the self-directed setting. Not only do they choose which points the learner can see, they can adaptively choose true labels in response to the learner's queries.
In this section, we consider a weaker variant of the realizable setting in which the adversary decides the true labels at the start of the game. 
This setting is colloquially called the \emph{oblivious} adversary setting.

For non-oblivious adversaries in the regime where $\VC(\H) < \infty$, Theorem \ref{thm:vc-apple-tasting} gives the deterministic algorithm \textsc{Prune-and-Eat} which achieves $O(\sqrt{\VC(\H)n \log n})$ mistakes. In this section we show that the $\sqrt{\log n}$ factor can be removed for oblivious adversaries and that the resulting bound is tight.

Since the true concept $h^*$ is now chosen at the start of the game, the most natural predictor is the one-inclusion graph of \cite{Haussler1994}. We take a sample of size $\sqrt{\VC(\H)n}$ from the provided set of points $S$ and treat it as training data to construct the one-inclusion graph. Using this predictor, we generalize to the other points in $S$.

\begin{algorithm}[H]
\caption{\textsc{Sample-and-Eat}$(\H,S)$}
\label{alg:sample-and-eat}
\begin{algorithmic}[1]
    \State Draw $R \subseteq S$ uniformly at random among all subsets of $S$ of size $\left\lceil \sqrt{\VC(\H) n} \right\rceil$
    \State $S \gets S \setminus R$
    \State Query the points in $R$ in any order and predict $\hat y=1$ for all these points
    \State Obtain $h^*|_R$ \Comment{Projection of the true concept onto $R$}
    \State Construct the one-inclusion graph $G_\H(R,h^*|_R)$
    \State For each $x \in S$: 
    \State \hspace{\algorithmicindent} Predict $\hat y = G_\H(R,h^*|_R;x)$ \Comment{the one-inclusion graph prediction} 
    \State \hspace{\algorithmicindent} Ignore any feedback
    \State \hspace{\algorithmicindent} $S \gets S \setminus \{x\}$
\end{algorithmic}
\end{algorithm}

\begin{theorem}
\label{thm:oblivious}
    Let $\H$ be a concept class with $\VC(\H)<\infty$ and assume the adversary is oblivious. Then $\H$ is learnable by algorithm \textnormal{\textsc{Sample-and-Eat}} with at most $2\sqrt{\VC(\H)n} + 1$ mistakes in expectation.
\end{theorem}
\begin{proof}
    The algorithm makes mistakes in two phases, which we bound separately:
    \begin{itemize}
        \item $M_1$: Mistakes made while predicting $1$ for all points in $R$. We can naively bound $M_1 \leq |R| \leq \sqrt{\VC(\H)n}+1$.
        \item $M_2$: Mistakes made by the one-inclusion graph predictor. We know the one-inclusion graph makes a mistake with probability at most $\frac{\VC(\H)}{|R|}$ on an unseen point.
        Therefore, $\E[M_2] \le |S \setminus R|\cdot \frac{\VC(\H)}{|R|} \le \sqrt{\VC(\H)n}$.
    \end{itemize} 
    
    Combining the two bounds, we get $\E[M_1+M_2] \leq 2\sqrt{\VC(\H)n}+1$ as desired.
\end{proof}

The upper bound of Theorem \ref{thm:oblivious} still has a $\sqrt{\VC(\H)}$ factor. We now show that in general, this factor cannot be removed. We use several independent copies of the structure from Lemma \ref{lemma:sddimlowerbound}.

\begin{theorem}
    \label{thm:vclowerbound}
    \raggedright{For all $d \in \mathbb{N}$, there exists a concept class $\H_d$ such that $\VC(\H_d) = d$ and  
    $\Mopt(\H_d, n) \in \Omega(\sqrt{n d})$, even against an oblivious adversary.}
    \label{theorem:vc-realizable-lower-bound}
\end{theorem}

\begin{proof}
    Let $\H_1$ be the infinite singleton concept class. That is, let $\X = \mathbb{N}$ be its instance space and include one concept $h_k$ for each $k \in \X$, where $h_k(k) = 1$ and $h_k(j) = 0$ for all $j \neq k$. We then let $\H_d$ consists of $d$ independent copies of $\H_1$. Since $\VC(\H_1) = 1$, we have $\VC(\H_d) = d$ as desired.

    We now show that $\Mopt(\H_d, n) \in \Omega(\sqrt{n d})$. Since $\Sdim(\H_1) = \infty$, by Lemma \ref{lemma:sddimlowerbound} for any $k$ we can construct instance spaces of size $k^2$ on which an oblivious adversary can force $\Omega(k)$ mistakes in expectation. Let $n \in \mathbb{N}$ be arbitrary. We partition $n$ into $d$ blocks of size $n/d$ each, then fill each block with the construction of Lemma \ref{lemma:sddimlowerbound} using the $d$ copies of $\H_1$ in $\H_d$. By playing independently on each block, the adversary can therefore force $\Omega(\sqrt{n/d})$ mistakes on all $d$ blocks, for a total of $\Omega(\sqrt{n d})$.
\end{proof}

\section{A class in the gap}
\label{sec:classingap}
Three of the four regimes that Theorem \ref{thm:dimensionforrates} classifies include common concept classes:
\begin{itemize}
    \item The class containing all concepts has infinite VC dimension;
    \item The infinite singletons concept class has finite VC dimension but infinite $\Sdim$; and
    \item Finite concept classes have finite $\Sdim$ and $\SDdim$.
\end{itemize}
The remaining regime has finite $\Sdim$ but infinite $\SDdim$. In this section, we show that it too is nonempty.
We use a combination of Lemma \ref{lemma:vc-star-equiv} and a construction of \cite{kuhlmann1999teachingintersectionclosed}, whose relevant details are summarized by the following proposition.

\begin{proposition}[Sub-claims of Theorem 2 from \cite{kuhlmann1999teachingintersectionclosed}]
    Let $\H$ be an intersection-closed concept class on an instance space $\X$. Suppose $2 \leq \VC(\H) < \infty$. Then there exists an intersection-closed concept class $\H'$ on an instance space $\X'$ such that:
    \begin{itemize}
        \item $\VC(\H') = \VC(\H)$, and
        \item For all $x \in \X'$ and $y \in \{0, 1\}$, the version space $\{h \in \H' \mid h(x) = y\}$ contains a copy of $\H$.
    \end{itemize}
    Furthermore, if $\X$ is finite, then one can take $|\X'| = (|\X|+1) \cdot |\X|$.
    \label{prop:kuhlmann-intersection-closed-construction}
\end{proposition}

\begin{theorem}
\label{thm:intersectionclosed}
    There exists a concept class $\H$ for which $\Sdim(\H) = 2$, $\SDdim(\H)=\infty$ and \\
    $\Mopt(\H, n) \in \Omega(\log \log n)$.
\end{theorem}

\begin{proof}
    We use the construction of Proposition \ref{prop:kuhlmann-intersection-closed-construction} recursively. It produces intersection-closed concept classes, but by reversing the labels we can produce union-closed concept classes instead. Then Lemma \ref{lemma:vc-star-equiv} allows us to relate their VC-dimension to their $\emptyset$-centered star number.

    Specifically, let $\X_0$ be a set of size $2$ and let $\H_0$ be the set of all concepts on $\X_0$. We then have $\VC(\H_0) = 2$ and that it is union-closed. Recursively apply the construction of Proposition \ref{prop:kuhlmann-intersection-closed-construction} but with inverted labels to produce instance spaces $\X_1, \X_2, \ldots$ and union-closed concept classes $\H_1, \H_2, \ldots$ that still satisfy $\VC(\H_i) = 2$ for all $i$. Each $\H_i$ contains copies of $\H_{i-1}$, so by identifying $\X_{i-1}$ with one of these copies we can turn this infinite sequence of concept classes into a single infinite concept class. We take this limiting concept class to be $\H$ and the limiting instance space to be $\X$.

    We first show that $\Sdim(\H) = 2$. Because $\H$ is union-closed, by Lemma \ref{lemma:vc-star-equiv} we have $\Sdim(\H) = \VC(\H)$. Since $\H$ contains a copy of $\H_0$ we have $\VC(\H) \geq 2$, and since any shattered set of size $3$ would be contained in some finite $\X_i$ we have $\VC(\H) \leq 2$. Therefore, $\Sdim(\H) = \VC(\H) = 2$.

    We now show that $\Mopt(\H, n) \in \Omega(\log \log n)$. As a stepping stone, we show by induction that $M(\H_i, \X_i) \geq i/2$, recalling that $M(\H_i, \X_i)$ denotes the expected number of mistakes an adversary can force when presenting the instance space $\X_i$ to the learner. The base case $M(\H_0, \X_0) \geq 0$ is trivial. For the inductive step, suppose $M(\H_i, \X_i) \geq i/2$ for some $i \geq 0$. When the learner is presented with $\X_{i+1}$, the adversary can choose the true label for the learner's first point uniformly at random. This forces half a mistake on average. By construction, there then exists a copy of $\H_i$ that agrees with this random label, so by the inductive hypothesis the adversary can force $i/2$ more mistakes on average. They can therefore force $(i+1)/2$ in total, completing the inductive step.

    For $\Mopt(\H, n)$, the adversary chooses the largest $\X_i$ such that $|\X_i| \leq n$ and uses their optimal strategy for $\H_i$. The learner then incurs $i/2$ mistakes on average. Since $|\X_{j+1}| = (|\X_j|+1) \cdot |\X_j|$ for all $j$, in terms of $n$ we have $i/2 \in \Omega(\log \log n)$. Since the adversary's strategy works even for full-information feedback, we have $\SDdim(\H) = \infty$ as well.
\end{proof}

\section{AI Disclosure}

We used GPT 5.6 Pro, GPT 6 Astra, and Opus 5.5 to produce some of our arguments. Our initial hypothesis was that there was a trichotomy defined by $\Sdim(\H)$ and $\VC(\H)$, and we spent a long time without AI assistance trying to show a constant upper bound for classes with finite $\Sdim(\H)$. This, of course, turned out to be incorrect, and prompting GPT 5.6 Pro directed us to the construction of \cite{kuhlmann1999teachingintersectionclosed} as the basis for a counterexample. Prompting with GPT 5.6 Pro and GPT 6 Astra provided the upper bounds of Theorem \ref{thm:upperboundlognconstant}. A more general prompting with Opus 5.5 produced the oblivious-adversary upper bound of Theorem \ref{thm:oblivious}, which was not a setting we were considering. All writing in this paper is our own.

\section{Conclusion and Future Work} 
We introduced self-directed online classification under apple-tasting feedback and determined its minimax rates in both the realizable and the agnostic settings. 
In the realizable setting, we obtain a tetrachotomy where the expected number of minimax mistakes can be $\Theta(1)$, $\omega(1)$ and $O(\log n)$, $\tilde \Theta(\sqrt{n})$, or $\Theta(n)$. We suspect, however, that the $\log$ factors in the $\omega(1) \cap O(\log n)$ and $\tilde \Theta(\sqrt{n})$ regimes might hide more fundamental characterizations.

In the $\tilde \Theta(\sqrt{n})$ regime, if we assume the adversary is oblivious then Theorem \ref{thm:oblivious} attains a mistake bound of $O(\sqrt{\VC(\H)n})$. We also obtain a matching lower bound attained by a specific class. 
Is the extra $\sqrt{\log n}$ factor in Theorem \ref{thm:vc-apple-tasting} an artifact of the adaptive adversary, the deterministic algorithm, or both? 
There are also classes with large $\VC$ dimension that do not have proportionally larger mistake bounds. 
For instance, a concept class consisting of the infinite singletons concept class plus a shattered set of size $d-1$ has VC dimension $d$ but behaves almost identically to the infinite singletons concept class. Is there a combinatorial dimension that captures this nuance?

\bibliography{biblography}

\begin{thebibliography}{29}
\providecommand{\natexlab}[1]{#1}
\providecommand{\url}[1]{\texttt{#1}}
\expandafter\ifx\csname urlstyle\endcsname\relax
  \providecommand{\doi}[1]{doi: #1}\else
  \providecommand{\doi}{doi: \begingroup \urlstyle{rm}\Url}\fi

\bibitem[Alon et~al.(2015)Alon, Cesa-Bianchi, Dekel, and Koren]{alon2015beyond}
Noga Alon, Nicol{\`o} Cesa-Bianchi, Ofer Dekel, and Tomer Koren.
\newblock Online learning with feedback graphs: Beyond bandits.
\newblock In \emph{Proceedings of The 28th Conference on Learning Theory}, volume~40, pages 23--35. PMLR, 2015.
\newblock URL \url{https://proceedings.mlr.press/v40/Alon15.html}.

\bibitem[Alweiss et~al.(2024)Alweiss, Huang, and Sellke]{alweiss2024improvedlowerboundfrankls}
Ryan Alweiss, Brice Huang, and Mark Sellke.
\newblock Improved lower bound for frankl's union-closed sets conjecture, 2024.
\newblock URL \url{https://arxiv.org/abs/2211.11731}.

\bibitem[Bart{\'o}k et~al.(2014)Bart{\'o}k, Foster, P{\'a}l, Rakhlin, and Szepesv{\'a}ri]{bartok2014partial}
G{\'a}bor Bart{\'o}k, Dean~P. Foster, D{\'a}vid P{\'a}l, Alexander Rakhlin, and Csaba Szepesv{\'a}ri.
\newblock Partial monitoring---classification, regret bounds, and algorithms.
\newblock \emph{Mathematics of Operations Research}, 39\penalty0 (4):\penalty0 967--997, 2014.
\newblock \doi{10.1287/moor.2014.0663}.

\bibitem[Ben-David et~al.(1995)Ben-David, Eiron, and Kushilevitz]{bendavid1995self}
Shai Ben-David, Nadav Eiron, and Eyal Kushilevitz.
\newblock On self-directed learning.
\newblock In \emph{Proceedings of the Eighth Annual Conference on Computational Learning Theory}, COLT '95, page 136–143, New York, NY, USA, 1995. Association for Computing Machinery.
\newblock ISBN 0897917235.
\newblock \doi{10.1145/225298.225314}.
\newblock URL \url{https://doi.org/10.1145/225298.225314}.

\bibitem[Bousquet et~al.(2020)Bousquet, Hanneke, Moran, and Zhivotovskiy]{bousquet2020proper}
Olivier Bousquet, Steve Hanneke, Shay Moran, and Nikita Zhivotovskiy.
\newblock Proper learning, helly number, and an optimal svm bound.
\newblock In \emph{Conference on Learning Theory}, pages 582--609. PMLR, 2020.

\bibitem[Bruhn and Schaudt(2013)]{bruhn2013journeyunionclosedsetsconjecture}
Henning Bruhn and Oliver Schaudt.
\newblock The journey of the union-closed sets conjecture, 2013.
\newblock URL \url{https://arxiv.org/abs/1309.3297}.

\bibitem[Cesa-Bianchi et~al.(2006)Cesa-Bianchi, Lugosi, et~al.]{cesa2006prediction}
Nicolo Cesa-Bianchi, G{\'a}bor Lugosi, et~al.
\newblock \emph{Prediction, learning, and games}, volume~1.
\newblock Cambridge university press Cambridge, 2006.

\bibitem[Cesa-Bianchi et~al.(2007)Cesa-Bianchi, Mansour, and Stoltz]{cesa2007improved}
Nicolo Cesa-Bianchi, Yishay Mansour, and Gilles Stoltz.
\newblock Improved second-order bounds for prediction with expert advice.
\newblock \emph{Machine Learning}, 66\penalty0 (2):\penalty0 321--352, 2007.

\bibitem[Chase and Lovett(2022)]{chase2022approximateunionclosedconjecture}
Zachary Chase and Shachar Lovett.
\newblock Approximate union closed conjecture, 2022.
\newblock URL \url{https://arxiv.org/abs/2211.11689}.

\bibitem[Chase and Mehalel(2025)]{chase2025deterministicappletasting}
Zachary Chase and Idan Mehalel.
\newblock Deterministic apple tasting, 2025.
\newblock URL \url{https://arxiv.org/abs/2410.10404}.

\bibitem[Devulapalli and Hanneke(2024)]{devulapalli2024dimensionselfdirectedlearning}
Pramith Devulapalli and Steve Hanneke.
\newblock The dimension of self-directed learning.
\newblock In \emph{Proceedings of the 35th International Conference on Algorithmic Learning Theory}, volume 237, pages 544--573, 2024.

\bibitem[Filmus et~al.(2025)Filmus, Hanneke, Mehalel, and Moran]{filmus2025optimalpredictionusingexpert}
Yuval Filmus, Steve Hanneke, Idan Mehalel, and Shay Moran.
\newblock Optimal prediction using expert advice and randomized littlestone dimension, 2025.
\newblock URL \url{https://arxiv.org/abs/2302.13849}.

\bibitem[Gilmer(2022)]{gilmer2022constantlowerboundunionclosed}
Justin Gilmer.
\newblock A constant lower bound for the union-closed sets conjecture, 2022.
\newblock URL \url{https://arxiv.org/abs/2211.09055}.

\bibitem[Goldman and Sloan(1994)]{goldman1994power}
Sally~A Goldman and Robert~H Sloan.
\newblock The power of self-directed learning.
\newblock \emph{Machine Learning}, 14\penalty0 (3):\penalty0 271--294, 1994.

\bibitem[Hanneke(2016)]{hanneke2016refined}
Steve Hanneke.
\newblock Refined error bounds for several learning algorithms.
\newblock \emph{Journal of Machine Learning Research}, 17\penalty0 (135):\penalty0 1--55, 2016.

\bibitem[Hanneke(2024)]{hanneke2024starnumbereluderdimension}
Steve Hanneke.
\newblock The star number and eluder dimension: Elementary observations about the dimensions of disagreement.
\newblock In \emph{Proceedings of the Thirty Seventh Conference on Learning Theory}, volume 247, pages 2308--2359, 2024.

\bibitem[Hanneke and Yang(2015)]{hanneke15a}
Steve Hanneke and Liu Yang.
\newblock Minimax analysis of active learning.
\newblock \emph{Journal of Machine Learning Research}, 16\penalty0 (109):\penalty0 3487--3602, 2015.
\newblock URL \url{http://jmlr.org/papers/v16/hanneke15a.html}.

\bibitem[Haussler et~al.(1994)Haussler, Littlestone, and Warmuth]{Haussler1994}
D.~Haussler, N.~Littlestone, and M.K. Warmuth.
\newblock Predicting {0, 1}-functions on randomly drawn points.
\newblock \emph{Information and Computation}, 115\penalty0 (2):\penalty0 248–292, December 1994.
\newblock ISSN 0890-5401.
\newblock \doi{10.1006/inco.1994.1097}.
\newblock URL \url{http://dx.doi.org/10.1006/inco.1994.1097}.

\bibitem[Helmbold et~al.(2000)Helmbold, Littlestone, and Long]{Helmbold2000}
David~P. Helmbold, Nicholas Littlestone, and Philip~M. Long.
\newblock Apple tasting.
\newblock \emph{Information and Computation}, 161\penalty0 (2):\penalty0 85–139, 2000.
\newblock ISSN 0890-5401.
\newblock \doi{10.1006/inco.2000.2870}.
\newblock URL \url{http://dx.doi.org/10.1006/inco.2000.2870}.

\bibitem[Kuhlmann(1999)]{kuhlmann1999teachingintersectionclosed}
Christian Kuhlmann.
\newblock On teaching and learning intersection-closed concept classes.
\newblock In Paul Fischer and Hans~Ulrich Simon, editors, \emph{Computational Learning Theory (EuroCOLT 1999)}, pages 168--182. Springer Berlin Heidelberg, 1999.

\bibitem[Liu(2023)]{liu2023improvinglowerboundunionclosed}
Jingbo Liu.
\newblock Improving the lower bound for the union-closed sets conjecture via conditionally iid coupling, 2023.
\newblock URL \url{https://arxiv.org/abs/2306.08824}.

\bibitem[Pebody(2022)]{pebody2022extensionmethodgilmer}
Luke Pebody.
\newblock Extension of a method of gilmer, 2022.
\newblock URL \url{https://arxiv.org/abs/2211.13139}.

\bibitem[Raman et~al.(2024)Raman, Subedi, Raman, and Tewari]{raman2024appletastingcombinatorialdimensions}
Vinod Raman, Unique Subedi, Ananth Raman, and Ambuj Tewari.
\newblock Apple tasting: Combinatorial dimensions and minimax rates.
\newblock In \emph{Proceedings of the Thirty Seventh Conference on Learning Theory}, volume 247, pages 4358--4380, 2024.

\bibitem[Sawin(2023)]{sawin2023improvedlowerboundunionclosed}
Will Sawin.
\newblock An improved lower bound for the union-closed set conjecture, 2023.
\newblock URL \url{https://arxiv.org/abs/2211.11504}.

\bibitem[Tsybakov(2004)]{tsybakov2004optimal}
Alexander~B Tsybakov.
\newblock Optimal aggregation of classifiers in statistical learning.
\newblock \emph{The Annals of Statistics}, 32\penalty0 (1):\penalty0 135--166, 2004.

\bibitem[Vapnik and Chervonenkis(1971)]{Vapnik1971}
V.~N. Vapnik and A.~Ya. Chervonenkis.
\newblock On the uniform convergence of relative frequencies of events to their probabilities.
\newblock \emph{Theory of Probability \& Its Applications}, 16\penalty0 (2):\penalty0 264–280, January 1971.
\newblock ISSN 1095-7219.
\newblock \doi{10.1137/1116025}.
\newblock URL \url{http://dx.doi.org/10.1137/1116025}.

\bibitem[Vovk(1995)]{vovk1995game}
Vladimir~G Vovk.
\newblock A game of prediction with expert advice.
\newblock In \emph{Proceedings of the eighth annual conference on Computational learning theory}, pages 51--60, 1995.

\bibitem[Vovk(1990)]{vovk1990aggregating}
Volodimir~G Vovk.
\newblock Aggregating strategies.
\newblock In \emph{Proceedings of the third annual workshop on Computational learning theory}, pages 371--386, 1990.

\bibitem[Wiener et~al.(2015)Wiener, Hanneke, and El-Yaniv]{wiener2015compression}
Yair Wiener, Steve Hanneke, and Ran El-Yaniv.
\newblock A compression technique for analyzing disagreement-based active learning.
\newblock \emph{The Journal of Machine Learning Research}, 16\penalty0 (1):\penalty0 713--745, 2015.

\end{thebibliography}

\appendix

\end{document}